# From Task Executors to Research Partners: Evaluating AI Co-Pilots Through Workflow Integration in Biomedical Research

Lukas Weidener*; Marko Brkić; Chiara Baccin; Mihailo Jovanović;
Emre Ulgac; Alex Dobrin; Johannes Weniger; Martin Vlas; Ritvik Singh; Aakaash Meduri

**Abstract**. Artificial intelligence systems are increasingly deployed in biomedical research. However, current evaluation frameworks may inadequately assess their effectiveness as research collaborators. This rapid review examines benchmarking practices for AI systems in preclinical biomedical research. Three major databases and two preprint servers were searched from January 1, 2018 to October 31, 2025, identifying 14 benchmarks that assess AI capabilities in literature understanding, experimental design, and hypothesis generation. The results revealed that all current benchmarks assess isolated component capabilities, including data analysis quality, hypothesis validity, and experimental protocol design. However, authentic research collaboration requires integrated workflows spanning multiple sessions, with contextual memory, adaptive dialogue, and constraint propagation. This gap implies that systems excelling on component benchmarks may fail as practical research co-pilots. A process-oriented evaluation framework is proposed that addresses four critical dimensions absent from current benchmarks: dialogue quality, workflow orchestration, session continuity, and researcher experience. These dimensions are essential for evaluating AI systems as research co-pilots rather than as isolated task executors.

# 1. Introduction

Artificial intelligence systems are increasingly deployed across biomedical research workflows, from literature synthesis and hypothesis generation to experimental design and data analysis (Wang & Barabási, 2021; Kitano, 2021). Recent advances in large language models and agentic AI architectures promise to transform how scientists conduct research, potentially accelerating discovery timelines and expanding the scope of tractable investigations (Achiam et al., 2023; Xi et al., 2025). Recursion Pharmaceuticals reports executing 2.2 million experiments weekly through AI-guided automation (Recursion, 2025), while laboratory automation markets project 7-8% annual growth driven by AI integration (MarketsandMarkets, 2024). These developments suggest a fundamental shift from AI as a narrow task executor to a collaborative research partner.

However, the deployment of AI systems in high-stakes research contexts requires rigorous evaluation frameworks that ensure scientific validity, reproducibility, and safety (Amodei et al., 2016; Sandbrink, 2023). Benchmarking, a systematic assessment of system capabilities against standardized tasks, has

*Corresponding author: lukas@bio.xyz

ORCID: L. Weidener (0000-0002-7132-8826); M. Brkić (0009-0008-5296-2697); M. Jovanović (0009-0007-1339-7544); C. Baccin (0000-0003-1251-6947); R. Singh (0009-0003-6612-6685); E. Ulgac (0009-0004-9382-8014); A. Dobrin (0009-0006-8957-3914); A. Meduri (0009-0001-1586-013X)

proven essential for tracking progress, enabling comparisons, and identifying limitations across AI applications (Bowman & Dahl, 2021; Ethayarajh & Jurafsky, 2020). Specialized benchmarks have emerged in the biomedical domain to evaluate the literature understanding (Gu et al., 2021; Nentidis et al., 2025), hypothesis generation (Tyagin & Safro, 2024), experimental design (Laurent et al., 2024; O'Donoghue et al., 2023), and closed-loop discovery (Roohani et al., 2024). These evaluation frameworks reflect a sophisticated understanding of domain-specific requirements, including biological knowledge, experimental reasoning, and safety critical decision-making.

A critical question remains largely unaddressed: Do current benchmarks evaluate what actually matters for effective AI-human research collaboration? Biomedical research differs fundamentally from isolated task execution. Scientists rarely complete projects in single sessions; instead, research progresses episodically over days or weeks, with irregular interaction patterns (Cetina, 1999; Latour & Woolgar, 2013). Budget constraints emerge as mid-level projects that require adaptive replanning. Experimental results necessitate hypothesis refinement through iterative dialogue, where collaborators must gracefully incorporate critique rather than defend conclusions (Collins, 1992). For example: A postdoc analyzing cardiac data might explore results Monday, refine hypotheses Tuesday, adapt protocols Thursday following budget discussions, and integrate everything into a proposal the following week. Existing benchmarks separately evaluate computational analysis quality (Miller et al., 2025), hypothesis validity (Tyagin & Safro, 2024), and laboratory protocol appropriateness (Laurent et al., 2024). None would assess whether the AI system remembered Tuesday's hypothesis refinement when designing Thursday's experiments, whether Thursday's budget constraint propagated to Monday's proposal recommendations, or whether the agent asked appropriate clarifying questions before committing to interpretation. An overview of a potential flow and the gaps of current assessments are shown in Figure 1.

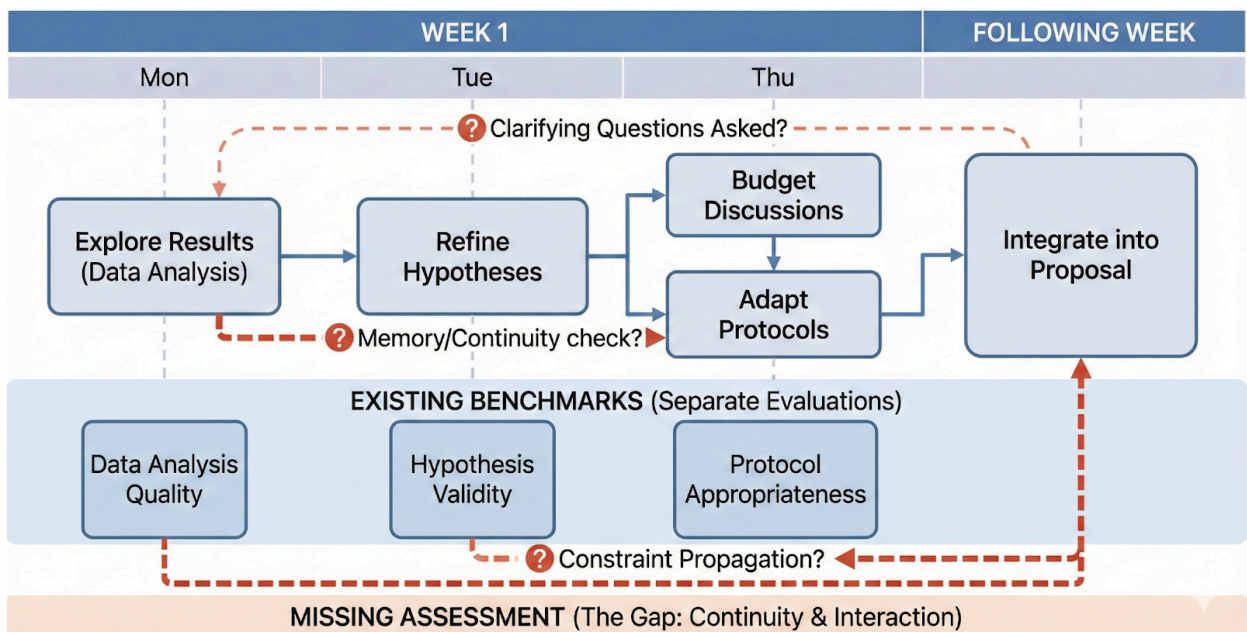

*Figure 1. Integrated research workflow versus isolated benchmark evaluation.* *Research workflows span multiple sessions across days or weeks, requiring memory continuity, constraint propagation, and appropriate dialogue. Existing benchmarks separately evaluate component capabilities (data analysis, hypothesis validity, protocol design) but do not assess the integration dimensions (red dashed arrows) critical for authentic collaboration.*

Moreover, evaluating AI systems primarily against human expert performance may fundamentally mischaracterize their potential value; AI systems might discover novel problem-solving approaches that bypass human cognitive constraints entirely, capabilities that human-benchmarked evaluation frameworks cannot capture (Eloundou et al., 2023; Korinek & Stiglitz, 2021). This review examines benchmarking practices for AI systems in preclinical biomedical research using a rapid review methodology (Tricco et al., 2015; Garritty et al., 2021). Beyond cataloging existing evaluation approaches, this study critically analyzes what current benchmarks measure, what capabilities determine effective research collaboration, and whether a gap exists between evaluation paradigms and authentic research workflows.

To address these questions, this review pursued three objectives:
- Synthesize existing benchmarking approaches for AI systems in pre-clinical biomedical research, systematically comparing what aspects are evaluated and how methodologies differ across task types

- Identify limitations in current evaluation paradigms and analyze gaps between isolated task assessment and integrated research collaboration requirements

- Explore and outline a process-oriented framework for evaluating AI research co-pilots in integrated research workflows

## 2. Methodology

This rapid review followed the established rapid review methodology, employing streamlined search and screening procedures to efficiently synthesize current benchmarking practices for AI systems in preclinical biomedical research (Tricco et al., 2015; Garritty et al., 2021). To balance comprehensiveness with time efficiency, the review employed focused search strategies, abbreviated timeframes, and pragmatic inclusion criteria, while maintaining methodological rigor appropriate for the research question.

Given the rapidly evolving nature of AI benchmarking in biomedical research, the search included limited gray literature sources focusing on major AI research institutions to capture recent evaluation frameworks that may not yet appear in peer-reviewed publications (Paez, 2017).

Definitions used in this review:
- **Biomedical Scientific AI:** Artificial intelligence systems designed to perform or assist with preclinical biomedical research tasks (Xi et al., 2025).This encompasses systems that integrate capabilities including: (1) large language models and AI agents for literature review, hypothesis generation, experimental design, protocol planning, and data analysis; (2) domain-specific models for protein structure prediction, drug discovery, or molecular property prediction; (3) multi-agent systems for collaborative biomedical research; and (4) AI-assisted research software with core scientific functions. Clinical diagnostic systems, patient care applications, educational tools, and business applications were excluded. For effective research co-pilots, these capabilities must be integrated rather than functioning as standalone components
- **Benchmarking/Evaluation:** A systematic evaluation framework with explicit protocols for assessing AI system performance on defined biomedical research tasks (Bowman & Dahl, 2021; Kiela et al., 2021). To qualify for inclusion, frameworks required: (1) publication in peer-reviewed venues, preprints, or substantive gray literature from research institutions; (2) explicit evaluation protocols specifying tasks, methodology, and performance metrics; and (3) reproducibility information including task descriptions and evaluation criteria. Datasets without evaluation protocols were excluded from analysis.
- **Pre-clinical Biomedical Research:** Research investigating biological mechanisms, disease processes, and therapeutic interventions prior to human clinical trials (Begley & Ioannidis, 2015). These include: (1) basic biological sciences (molecular biology, genomics, systems biology, immunology, and neuroscience); (2) drug discovery and development (target identification, lead optimization, pharmacology); (3) translational research (disease modeling, biomarker discovery, mechanism studies), and (4) biomedical informatics (bioinformatics, computational biology, omics analysis). Encompasses both computational sciences and wet-lab experimental sciences. It excludes clinical medicine, epidemiology, health services research, and public health applications.

## 2.1 Search strategy

Three core databases were systematically searched: PubMed/MEDLINE, Web of Science, and IEEE Xplore. Two preprint servers were searched: bioRxiv and arXiv (cs.AI, cs.CL, and cs.LG sections). Gray literature sources included technical reports from major AI research laboratories with established biomedical AI programs (such as OpenAI, Google DeepMind, Anthropic, Allen Institute for AI, and FutureHouse). Forward citation tracking of the included articles was performed to ensure that key benchmarks were not missed.

The timeframe was from January 1, 2018, to October 31, 2025, with the starting point marking when transformer architectures (Vaswani et al., 2017) established the prerequisite capabilities for contemporary scientific AI applications. The focus was on benchmarking and evaluation frameworks for AI systems applied to pre-clinical biomedical research, defined as artificial intelligence designed to perform, assist with, or automate tasks within the biomedical research process from literature analysis through experimental design and data interpretation. Boolean operators ("AND," "OR," "NOT") were employed to refine and specify search results across databases. The search used the following keywords organized by key concepts:

- **AI systems and technologies:** "Artificial intelligence," "Large language model," "AI agent," "Agentic AI," "Biomedical AI," specific model names (GPT, BERT, Claude, BioBERT), "Retrieval-augmented generation," among others;
- **Benchmarking and evaluation:** "Benchmark," "Evaluation framework," "Performance evaluation," "LLM-as-judge," "Model-graded evaluation," "Safety evaluation," "Protocol validation," among others;

- **Biomedical research tasks and workflows:** "Literature review," "Hypothesis generation," "Experimental design," "Protocol planning," "Data analysis," "Scientific reasoning," "Drug discovery," "CRISPR design," "Gene editing," among others;
- **Biomedical domains and applications:** "Biomedical research," "Preclinical research," "Molecular biology," "Genomics," "Bioinformatics," "Drug development," "Protein design," "Disease modeling," among others."

Search strategies were translated across database-specific syntax requirements using the Polyglot Search Translator (Clark et al., 2020). When databases did not support full advanced search syntax, multiple shorter keyword combinations were executed to maintain search comprehensiveness.

## 2.2 Study selection

Following the database searches, all records were imported into Rayyan (Ouzzani et al., 2016), a web-based systematic review tool for deduplication and screening. Following the Responsible AI in Evidence Synthesis (RAISE) guidance (Thomas et al., 2024), title and abstract screening was assisted by Rayyan; however, all inclusion and exclusion decisions were subject to human review. The screening proceeded in two stages. First, titles and abstracts were reviewed to align with the objective of mapping benchmarking practices for AI systems in preclinical biomedical research. Second, potentially eligible items underwent full-text assessment using predefined criteria conducted entirely by human reviewers without AI assistance. The included studies were exported to Zotero for reference management and citation formatting.

- The inclusion criteria were as follows: (i) biomedical research AI focus: AI systems designed for preclinical biomedical research tasks are the primary subject rather than general-purpose AI or clinical applications; (ii) benchmarking/evaluation content: explicit evaluation protocols, performance metrics, or assessment frameworks for biomedical scientific AI capabilities; (iii) preclinical biomedical domain application: basic biological sciences, drug discovery, translational research, or biomedical informatics; and (iv) quality standards for preprints and gray literature, documented evaluation methodology, explicit performance metrics or assessment criteria, and appropriate citation of sources or sufficient technical documentation.

- Exclusion criteria were: published before January 1, 2018; non-English language; no accessible full text; lack of thematic relevance (e.g., general AI benchmarks without biomedical application, clinical diagnostic/treatment systems, patient care applications, epidemiological studies, educational tools, or business applications); opinion pieces without methodological rigor; and preprints or gray literature lacking documented evaluation protocols, performance data, or appropriate technical documentation.

To enhance completeness, backward and forward citation tracking were performed on the included papers, with newly identified records subjected to the same screening pipeline.

## 2.3 Data extraction

Data collection and analysis were conducted using Google Sheets. For each record, the extracted fields included bibliographic information (title, year, authors, and publication type), benchmark characteristics (name, biomedical domain, and target AI system type), and evaluation design (aspects assessed, methodology, metrics employed, and task types covered).

## 2.4 Ethical considerations

As this review involved the analysis of publicly available published and gray literature without the collection of primary data or involvement of human participants, ethical approval was not required. All the included sources were appropriately cited according to academic standards.

# 3. Results

The search focused on three electronic databases (PubMed/MEDLINE, Web of Science, and IEEE Xplore) and two preprint servers (bioRxiv and arXiv) and yielded 3,247 records. Twelve additional records were identified through forward citation tracking and targeted gray literature searches of major AI research laboratories. After removing 1,456 duplicates, 1,803 unique records were retained for title and abstract screening, resulting in the exclusion of 1,628 records. Of the 175 publications retrieved for full-text review, 161 were excluded for the following primary reasons: lack of a systematic evaluation framework (n = 92), focus on clinical applications (n = 38), and insufficient alignment with preclinical biomedical scope (n = 31). Consequently, 14 publications met all inclusion criteria and were included in the final analysis.

## 3.1 Study characteristics

The 14 included benchmarks span from 2018 to 2025, with notable acceleration in recent years: two benchmarks were introduced in 2025, seven in 2024, two in 2023, one in 2021, and two in 2019. The concentration of the seven benchmarks (50%) in 2024 alone reflects the rapid evolution of AI capabilities for biomedical research applications.

Experimental design and protocol planning have emerged as a major focus in 2024. LAB-Bench evaluates language agents across eight categories spanning literature understanding to molecular cloning workflows, including 41 "human-hard" cloning scenarios that require multi-step reasoning (Laurent et al., 2024). The BioLP-bench introduces a protocol validation methodology by injecting critical mistakes into laboratory protocols to test whether LLMs can identify failure-causing errors (Ivanov, 2024). CRISPR-GPT provides an automated agent for CRISPR experimental design with 288 test cases covering experimental planning, sgRNA design, delivery methods, and gene-editing Q&A (Qu et al., 2024), whereas BioPlanner evaluates LLMs' ability to reconstruct laboratory protocols from high-level descriptions using pseudocode representations (O'Donoghue et al., 2023).

Hypothesis generation and closed-loop discovery frameworks include Dyport, which benchmarks hypothesis generation systems using dynamic knowledge graphs with temporal cutoff dates and importance-based scoring (Tyagin and Safro, 2024), and BioDiscoveryAgent, which designs genetic perturbation experiments through iterative closed-loop hypothesis testing, achieving a 21% improvement over the Bayesian optimization baselines (Roohani et al., 2024). The BioML-bench evaluates AI agents on end-to-end machine learning workflows across protein engineering, single-cell omics, biomedical imaging, and drug discovery (Miller et al., 2025).

Literature understanding and knowledge assessment constitute the most established categories. Foundational benchmarks include BLUE with 10 corpora across five tasks for biomedical language understanding (Peng et al., 2019), BLURB covering 13 datasets and six tasks with macro-average scoring (Gu et al., 2021), and BioASQ, the longest-running challenge (2012-2025) for large-scale biomedical semantic indexing and question answering (Nentidis et al., 2025). More specialized frameworks include PubMedQA for biomedical research question answering (Jin et al., 2019), ScholarQABench with 1,451 biomedical questions from PhD experts requiring multi-paper synthesis (Asai et al., 2024), and BioLaySumm focusing on the lay summarization of biomedical literature (Goldsack et al., 2024). Bioinfo-Bench assesses academic knowledge and data-mining capabilities, specifically for bioinformatic applications (Chen and Deng, 2023). Table 1 summarizes the study characteristics of the 14 benchmarks and related publications.

*Table 1. Characteristics of the identified publications*

| Title | Year | Authors | Publication Type | Benchmark Name |
|---|---|---|---|---|
| Language Agent Biology Benchmark | 2024 | Laurent et al. | Preprint (arXiv) | LAB-Bench |
| BioLP-bench: Evaluating Large Language Models on Biomedical Laboratory Protocols | 2024 | Ivanov | Preprint (bioRxiv) | BioLP-bench |
| CRISPR-GPT: An LLM Agent for Automated Design of Gene-Editing Experiments | 2024 | Qu et al., | Peer-reviewed (nature biomedical engineering) | CRISPR-GPT |
| BioPlanner: Automatic Evaluation of LLMs on Protocol Planning in Biology | 2023 | O'Donoghue et al. | Preprint (arXiv) | BioPlanner |

| Title | Year | Authors | Venue | Benchmark |
|---|---|---|---|---|
| Dyport: Dynamic Importance-Based Biomedical Hypothesis Generation Benchmarking Technique | 2024 | Tyagin & Safro | Peer-reviewed (BMC Bioinformatics) | Dyport |
| BioDiscoveryAgent: An AI Agent for Designing Genetic Perturbation Experiments | 2024 | Roohani et al. | Preprint (arXiv) | BioDiscoveryAgent |
| BioML-bench: Evaluating AI Agents on End-to-End Biomedical ML Tasks | 2025 | Miller et al. | Preprint (bioRxiv) | BioML-bench |
| Transfer Learning in Biomedical Natural Language Processing: An Evaluation of BERT and ELMo on Ten Benchmarking Datasets | 2019 | Peng et al. | Proceedings (BioNLP Workshop) | BLUE |
| Domain-Specific Language Model Pretraining for Biomedical Natural Language Processing | 2021 | Gu et al. | Peer-reviewed (ACM Transactions on Computing for Healthcare) | BLURB |
| Overview of BioASQ 2025: The thirteenth BioASQ challenge on large-scale biomedical semantic indexing and question answering | 2025 | Nentidis et al. | Conference proceedings (International Conference of the CLEF Association) | BioASQ |
| PubMedQA: A Dataset for Biomedical Research Question Answering | 2019 | Jin et al. | Conference proceedings (Empirical Methods in Natural Language Processing) | PubMedQA |
| ScholarQABench | 2024 | Asai et al. | Preprint (arXiv) | ScholarQABench |
| Bioinfo-Bench: A Simple Benchmark Framework for LLM Bioinformatics Skills Evaluation | 2023 | Chen & Deng | Preprint (bioRxiv) | Bioinfo-Bench |

| Overview of the BioLaySumm 2024 Shared Task on the Lay Summarization of Biomedical Research Articles | 2024 | Goldsack et al. | Proceedings (23rd Workshop on Biomedical Natural Language Processing) | BioLaySumm |

## 3.2 Evaluation Aspects Across Biomedical AI Benchmarks

Of the 14 included benchmarks, five primary evaluation dimensions emerged: traditional performance metrics, multistep reasoning and experimental planning, safety and error detection, knowledge synthesis and discovery, and tool-augmented workflows. These dimensions reflect the field's evolution from basic language understanding to complex research-assistance capabilities.

Traditional performance metrics form the foundation of evaluations across literature-understanding benchmarks. BLUE employs standard NLP metrics, including precision, recall, and F1-score, across five tasks spanning named entity recognition, relation extraction, and document classification (Peng et al., 2019). BLURB extends this approach with macro-average scoring across 13 datasets and six tasks, establishing model-agnostic leaderboards to compare the system performance (Gu et al., 2021). The BioASQ evaluates biomedical question answering through multiple phases, measuring article retrieval precision, snippet extraction accuracy, and answer quality against gold standard expert annotations (Nentidis et al., 2025). PubMedQA focuses specifically on yes/no/maybe answer accuracy for biomedical research questions (Jin et al., 2019), whereas Bioinfo-Bench assesses knowledge acquisition, analysis, and application capabilities across bioinformatics domains (Chen & Deng, 2023). BioLaySumm evaluates lay summarization quality using ROUGE scores, BERTScore, and readability metrics, including the Flesch-Kincaid Grade Level (Goldsack et al., 2024).

Multi-step reasoning and experimental planning capabilities represent critical advancements in the evaluation methodology. LAB-Bench distinguishes between tool-dependent and tool-free tasks across eight categories, with 41 "human-hard" cloning scenarios requiring trained molecular biologists over 10 minutes to solve, revealing that only Claude 3.5 Sonnet approaches human performance on certain tasks (Laurent et al., 2024). BioPlanner evaluates protocol planning by testing the ability of LLMs to reconstruct pseudocode representations of laboratory protocols from high-level descriptions, successfully demonstrating laboratory execution of a generated protocol (O'Donoghue et al., 2023). CRISPR-GPT assesses automated experimental design across 288 test cases covering sgRNA design, delivery method selection, and assay design, operating in three modes (Meta, Auto, Q&A) to evaluate different levels of automation (Qu et al., 2024). BioML-bench evaluates end-to-end machine learning workflows where agents must parse task descriptions, build pipelines, implement models, and submit predictions graded by established metrics, such as AUROC and Spearman correlation, finding that all evaluated agents underperform human baselines (Miller et al., 2025).

Safety and error detection are essential evaluation criteria for systems approaching laboratory deployment. BioLP-bench introduces a novel protocol validation methodology that injects critical mistakes into published laboratory protocols and evaluates whether LLMs can identify failure-causing errors, revealing that state-of-the-art language models demonstrate poor performance compared with human experts (Ivanov, 2024). The benchmark employs a model-graded evaluation requiring GPT-4o-level capabilities to avoid false positives and to establish rigorous standards for protocol understanding assessment. CRISPR-GPT incorporates ethical guardrails that prevent unethical applications, such as editing viruses or human embryos, demonstrating a safety-conscious benchmark design for dual-use technologies (Qu et al., 2024). The LAB-Bench includes a canary string for training contamination monitoring, addressing data leakage concerns that could artificially inflate the performance metrics (Laurent et al., 2024).

Knowledge synthesis and hypothesis generation evaluation extend beyond factual recall to assess discovery potential. Dyport benchmarks hypothesis generation systems using dynamic knowledge graphs with temporal cutoff dates and testing systems under realistic conditions while quantifying discovery importance through methods that assess not only accuracy but also potential impact in biomedical research (Tyagin & Safro, 2024). ScholarQABench evaluates multi-paper synthesis with 1,451 biomedical questions from PhD experts requiring integration across multiple sources, revealing that traditional models such as GPT-4o had near-zero citation F1 scores and hallucinated over 90% of the cited papers (Asai et al., 2024). BioDiscoveryAgent assesses closed-loop hypothesis testing for genetic perturbations by measuring both prediction accuracy and interpretability, achieving an average improvement of 21% over Bayesian optimization baselines while maintaining transparent decision making at every stage (Roohani et al., 2024).

Tool-augmented workflows and practical integration distinguish systems that are capable of leveraging external resources. LAB-Bench documentation shows that tool augmentation improves the performance of database queries, sequence reasoning, literature queries, and supplementary material analysis, with human expert baselines varying between tool and no-tool conditions (Laurent et al., 2024). BioDiscoveryAgent evaluates tool use, including biomedical literature search, code execution for biological dataset analysis, and AI critique for prediction evaluation, demonstrating the importance of multitool integration for research tasks (Roohani et al., 2024). BioML-bench requires agents to autonomously build complete machine-learning pipelines, including data preprocessing, model selection, hyperparameter tuning, and result submission, to assess whether biomedical specialization confers advantages over general-purpose architectures (Miller et al., 2025).

*Table 2. Evaluation aspects across biomedical AI benchmarks*

| Evaluation Dimension | Specifications |
| --- | --- |

| Traditional Performance Metrics | - **BLUE**: Uses standard NLP metrics across named entity recognition, relation extraction, and document classification (Peng et al., 2019).
- **BLURB**: Implements macro-average scoring across 13 datasets with model-agnostic leaderboards (Gu et al., 2021).
- **BioASQ**: Assesses article retrieval, snippet extraction, and answer quality against expert annotations (Nentidis et al., 2025).
- **PubMedQA**: Evaluates yes/no/maybe answer accuracy for biomedical research questions (Jin et al., 2019).
- **BioLaySumm**: Evaluates summarization quality using ROUGE, BERTScore, and readability metrics (Goldsack et al., 2024).
- **Bioinfo-Bench**: Measures knowledge acquisition, analysis, and application capabilities across bioinformatics domains (Chen & Deng, 2023). |
|---|---|
| Multi-Step Reasoning and Experimental Planning | - **LAB-Bench**: Distinguishes tool-dependent and tool-free tasks with 41 "human-hard" scenarios requiring 10+ minutes to solve (Laurent et al., 2024).
- **BioPlanner**: Evaluates protocol planning through pseudocode reconstruction validated by successful laboratory execution (O'Donoghue et al., 2023).
- **CRISPR-GPT**: Assesses automated experimental design across 288 test cases in multiple automation modes (Qu et al., 2024).
- **BioML-bench**: Tests end-to-end ML workflows requiring pipeline building and prediction submission graded by AUROC (Miller et al., 2025). |
| Safety and Error Detection | - **BioLP-bench**: Injects critical mistakes into protocols revealing poor LLM performance compared to human experts (Ivanov, 2024).
- **CRISPR-GPT**: Incorporates ethical guardrails preventing unethical applications like virus or embryo editing (Qu et al., 2024).
- **LAB-Bench**: Implements canary strings for contamination monitoring to address data leakage concerns (Laurent et al., 2024). |

| Knowledge Synthesis and Discovery | - **Dyport**: Benchmarks hypothesis generation using dynamic knowledge graphs quantifying discovery importance beyond accuracy (Tyagin & Safro, 2024).<br>- **ScholarQABench**: Evaluates multi-paper synthesis with 1,451 questions revealing 90%+ hallucination rates in traditional models (Asai et al., 2024).<br>- **BioDiscoveryAgent**: Assesses closed-loop hypothesis testing achieving 21% improvement over Bayesian optimization with transparent decision-making (Roohani et al., 2024). |
|---|---|
| Tool-Augmented Workflows | - **LAB-Bench**: Demonstrates significant performance improvements through tool augmentation for database queries and literature analysis (Laurent et al., 2024).<br>- **BioDiscoveryAgent**: Evaluates multi-tool integration including literature search, code execution, and AI critique (Roohani et al., 2024).<br>- **BioML-bench**: Requires autonomous ML pipeline building to assess biomedical specialization advantages over general architectures (Miller et al., 2025). |

## 3.3 Methodological Approaches Across Biomedical Task Types

Within the 14 benchmarks included, three unique categories of tasks were identified, each with its own methodological characteristics: understanding the literature and assessing knowledge, planning experimental design and protocols, and generating hypotheses along with closed-loop discovery. Each category exhibits distinct approaches to data construction, evaluation methodology, and performance measurement, reflecting the varying complexities and validation requirements of different research activities.

Literature understanding and knowledge assessment benchmarks predominantly employ curated datasets with expert-annotated ground-truth and automated evaluation metrics. Benchmarks with comprehensive expert annotations include ScholarQABench, CRISPR-GPT, LAB-Bench, and BioASQ Phase B, where domain experts generate both questions and reference answers, whereas PubMedQA and BioLaySumm incorporate expert-curated subsets alongside larger automated datasets. BLUE, BLURB, and Bioinfo-Bench utilize expert-annotated corpora but rely primarily on automated metrics such as F1-score, precision, and recall for evaluation. BioPlanner, Dyport, and BioMLbench predominantly employ automated evaluation through metrics including BLEU, SciBERTscore, Levenshtein distance, and ROC AUC, with BioMLbench validated against expert-populated leaderboards rather than crowd-sourced competitions. BLUE assembles 10 publicly available corpora across five tasks: named entity recognition, relation extraction, sentence similarity, document classification, and inference, utilizing standard NLP metrics (precision, recall, and F1-score) that enable automated, reproducible evaluation (Peng et al., 2019). BLURB extends this approach with 13 datasets spanning six tasks, introducing macro-average scoring across all tasks as the primary evaluation metric and establishing model-agnostic leaderboards accessible through HuggingFace (Gu et al., 2021). BioASQ represents the most comprehensive methodology with 13 annual editions (2013-2025), employing a two-phase evaluation, where Phase A assesses document and snippet retrieval from PubMed, and Phase B evaluates exact and ideal answers against expert-generated gold standards, with evaluation measures including mean precision, recall, F1-score, and mean average precision (Nentidis et al., 2025). PubMedQA provides three distinct subsets: PQA-L with 1,000 expert annotations, PQA-U with 61,200 unlabeled pairs, and PQA-A with 211,300 artificially generated instances, enabling the evaluation of models under different training regimes (Jin et al., 2019). Bioinfo-Bench evaluates bioinformatics skills across the knowledge acquisition, analysis, and application domains, providing a framework specifically targeting computational biology competencies (Chen and Deng, 2023). ScholarQABench introduced a critical innovation by evaluating citation accuracy, revealing that traditional models hallucinate over 90% of cited papers, while retrieval-augmented systems such as OpenScholar-8B achieved dramatic improvements in citation F1 scores (Asai et al., 2024). BioLaySumm employs a hybrid evaluation approach combining automated metrics (ROUGE, BERTScore) with readability measures (Flesch-Kincaid Grade Level, Dale-Chall Readability Score, Coleman-Liau Index) and alignment scores to address the multifaceted nature of lay summarization quality (Goldsack et al., 2024). This task category benefits from scalability through automation but faces challenges in capturing nuanced understanding beyond surface-level pattern matching.

Experimental design and protocol planning benchmarks introduce fundamentally different methodological challenges that require the evaluation of multistep reasoning, procedural correctness, and safety-critical decision-making. The LAB-Bench employs a comprehensive 8-category framework with 2,457 evaluation questions across 31 subtasks, distinguishing between tool-dependent and tool-free conditions to assess both reasoning capabilities and practical tool integration (Laurent et al., 2024). The benchmark uses an 80% public/20% private test split to prevent overfitting and training contamination monitoring. Critically, human expert baselines are established for each task, revealing that only specific models approach human performance on certain subtasks, while struggling dramatically on others. BioLP-bench introduces an error injection paradigm, deliberately inserting critical mistakes into published laboratory protocols and evaluating whether LLMs can identify failure-causing errors (Ivanov, 2024). This methodology employs a model-graded evaluation using the UK AI Safety Institute's Inspect framework, which requires GPT-4o-level scoring capabilities to distinguish between critical errors and benign variations (UK AI Safety Institute, 2024). The approach addresses a key limitation of traditional benchmarks: the inability to assess whether systems understand why protocol steps matter, rather than merely reproducing procedural text. BioPlanner converts natural language protocols into pseudocode representations using predefined laboratory actions and then evaluates the LLMs' ability to reconstruct the pseudocode from high-level descriptions (O'Donoghue et al., 2023). External validation through successful laboratory execution of a generated protocol provides ground-truth verification beyond the computational metrics. CRISPR-GPT employs a domain-specific benchmark with 288 test cases spanning experimental planning, sgRNA design, delivery methods, and assay design, incorporating safety guardrails that refuse unethical applications, such as virus editing or human embryo modification (Qu et al., 2024). The benchmark operates in three modes (Meta, Auto, Q&A), enabling evaluation of different automation levels and user interaction patterns. This task category necessitates expert validation and real-world execution testing, substantially limiting scalability but providing a high-confidence assessment of practical capabilities.

Hypothesis generation and closed-loop discovery benchmarks employ temporal validation and iterative experimental design methodologies that fundamentally differ from static knowledge assessments. Dyport introduced dynamic importance-based benchmarking using temporal knowledge graphs with cutoff dates, testing hypothesis generation systems under realistic conditions where future discoveries are unknown at the time of prediction (Tyagin & Safro, 2024). The methodology integrates knowledge from curated databases (creating a dynamic graph representation) with a novel method to quantify discovery importance, assessing not only whether predicted hypotheses prove correct but also their potential impact in biomedical research. The evaluation uses ROC AUC curves for different semantic type pairs (gene-gene, drug-disease), extending traditional link prediction benchmarks with impact-weighted metrics. BioDiscoveryAgent evaluates a closed-loop experimental design for genetic perturbations using the GeneDisco benchmark, where the agent iteratively designs experiments based on previous results (Roohani et al., 2024). The methodology employs Claude 3.5 Sonnet without training specialized machine learning models, achieving an average improvement of 21% over Bayesian optimization baselines and a 46% improvement on nonessential gene perturbation tasks. Critically, the evaluation includes one unpublished dataset, not in the language model's training data, providing a rigorous assessment of generalization capabilities. The system's interpretability, generating explicit biological reasoning for experimental choices, addresses the key requirement that scientists must understand and trust AI-generated hypotheses. BioML-bench extends closed-loop evaluation to complete machine learning workflows, requiring agents to parse task descriptions, autonomously build pipelines, implement models, and submit predictions graded by established metrics (AUROC and Spearman correlation) from existing biomedical ML competitions (Miller et al., 2025). Evaluation across four domains (protein engineering, single-cell omics, biomedical imaging, and drug discovery) with comparison against expert-populated leaderboards reveals that all evaluated agents underperform human baselines, and surprisingly, biomedical specialization does not confer consistent advantages over general-purpose architectures. This task category uniquely enables prospective validation through temporal hold-out or real-world experimental execution, providing the strongest evidence of genuine discovery capability, but requiring substantial computational resources and extended evaluation timelines.

Cross-cutting methodological innovations appeared across task categories. Model-graded evaluation has emerged in multiple benchmarks (BioLP-bench, ScholarQABench) as a scalable alternative to human expert annotation, although it introduces dependency on evaluator model quality and potential biases. Tool-augmented evaluation frameworks (LAB-Bench and BioDiscoveryAgent) distinguish systems capable of leveraging external resources from those relying solely on parametric knowledge, revealing that tool access substantially improves the performance of database queries, code execution, and literature search tasks. Contamination monitoring through canary strings (LAB-Bench) and unpublished test datasets (BioDiscoveryAgent) addresses the critical challenge of training data leakage in the era of web-scale pretraining. Hybrid evaluation combining automated metrics with expert assessment (BioASQ, BioLP-bench, and BioPlanner) balances scalability with validation rigor, enabling large-scale testing while maintaining quality control through selective human evaluation. The progression from static knowledge assessment to dynamic experimental design evaluation reflects the field's maturation from passive information retrieval to active research assistance, with corresponding increases in the methodological complexity and validation requirements.

*Table 3. Methodological approaches across biomedical task types*

| Task Category | Methodological Specifications |
|---|---|
| Literature Understanding and Knowledge Assessment | <ul><li>BLUE and BLURB employ expert-annotated datasets with automated NLP metrics (precision, recall, F1-score), with BLURB introducing macro-average scoring across 13 datasets and model-agnostic HuggingFace leaderboards (Peng et al., 2019; Gu et al., 2021).</li><li>BioASQ implements two-phase evaluation across 13 annual editions: Phase A for document retrieval and Phase B for answer validation against expert-generated gold standards using mean precision, recall, and F1-score (Nentidis et al., 2025).</li><li>PubMedQA provides three subsets (1,000 expert-annotated, 61,200 unlabeled, 211,300 artificially generated) enabling evaluation under different training regimes (Jin et al., 2019).</li><li>Bioinfo-Bench evaluates bioinformatics skills across knowledge acquisition, analysis, and application domains (Chen & Deng, 2023).</li><li>ScholarQABench introduces citation accuracy evaluation, revealing over 90% hallucination rates in traditional models compared to retrieval-augmented systems (Asai et al., 2024).</li><li>BioLaySumm combines automated metrics (ROUGE, BERTScore) with readability measures (Flesch-Kincaid, Dale-Chall, Coleman-Liau) and alignment scores for multi-faceted summarization assessment (Goldsack et al., 2024).</li></ul> |
| Experimental Design and Protocol Planning | <ul><li>LAB-Bench employs 2,457 questions across 31 subtasks with tool-dependent and tool-free conditions, incorporating contamination monitoring through canary strings (Laurent et al., 2024).</li><li>BioLP-bench injects critical errors into published protocols using model-graded evaluation via the Inspect framework (Ivanov, 2024).</li><li>BioPlanner converts protocols into pseudocode representations validated through successful laboratory execution (O'Donoghue et al., 2023).</li><li>CRISPR-GPT uses 288 expert-curated test cases across gene-editing planning, gRNA design, and delivery selection with consensus-based ground truth in three operational modes (Qu et al., 2024).</li></ul> |
| Hypothesis Generation and Closed-Loop Discovery | <ul><li>Dyport employs temporal knowledge graphs with impact-weighted metrics using ROC AUC for semantic type pair evaluation (Tyagin & Safro, 2024).</li><li>BioDiscoveryAgent evaluates iterative experimental design achieving 21% improvement over Bayesian optimization, validated on unpublished datasets to prevent contamination (Roohani et al., 2024).</li><li>BioML-bench requires autonomous ML pipeline building across four biomedical domains (drug discovery, imaging, protein engineering,</li></ul> |

| | single-cell omics), revealing agent underperformance against expert-populated leaderboard baselines (Miller et al., 2025). |
|---|---|
| Cross-Cutting Methodological Innovations | ● Model-graded evaluation as a scalable alternative to expert annotation (BioLP-bench, ScholarQABench) (Ivanov, 2024; Asai et al., 2024).<br>● Tool-augmented frameworks reveal substantial performance improvements on database queries and literature search (LAB-Bench, BioDiscoveryAgent) (Laurent et al., 2024; Roohani et al., 2024).<br>● Contamination monitoring through canary strings and unpublished datasets addressing training data leakage (LAB-Bench, BioDiscoveryAgent) (Laurent et al., 2024; Roohani et al., 2024).<br>● Hybrid evaluation combining automated metrics with expert assessment balances scalability with validation rigor (BioASQ, BioLP-bench, BioPlanner) (Nentidis et al., 2025; Ivanov, 2024; O'Donoghue et al., 2023). |

# 4. Discussion

This rapid review identified 14 benchmarks for evaluating AI systems in preclinical biomedical research, using streamlined systematic methods. While the focused search strategy and abbreviated timeframe prioritized efficiency, the resulting evidence base proved sufficient to identify the critical paradigm gap driving this analysis: current benchmarks evaluate isolated component capabilities, whereas authentic research collaboration requires integrated workflows and iterative human-AI feedback loops. None assess whether systems incorporate critique, adapt outputs based on researcher input, or improve through dialogue rather than defending initial conclusions. The convergence of findings across multiple benchmarks strengthens the confidence that this limitation represents a fundamental field-wide challenge, rather than an artifact of incomplete evidence synthesis.

## 4.1 Evolution of Evaluation Dimensions: From Metrics to Capabilities

The five evaluation dimensions identified were traditional performance metrics, multistep reasoning, safety and error detection, knowledge synthesis, and tool-augmented workflows, reflecting a fundamental shift in how the field conceptualizes AI competence for scientific applications. Traditional NLP metrics (precision, recall, and F1-score) that dominate literature understanding benchmarks emerge from information retrieval traditions, where task boundaries are well-defined and success criteria are unambiguous (Manning, 2009). However, these metrics correlate poorly with downstream research utility, which is a limitation extensively documented in NLP evaluation research (Lyu et al., 2024; Reiter, 2018). The "citation hallucination" crisis revealed by ScholarQABench, where GPT-4 fabricated over 90% of scientific references, exemplifies how systems can achieve impressive scores on surface-level metrics while lacking genuine understanding (Maynez et al., 2020; Ji et al., 2023).

The emergence of multistep reasoning evaluation responds to the recognized limitations of single-turn question answering in capturing research capabilities. LAB-Bench introduces 'human-hard' problems that require trained molecular biologists more than 10 minutes to solve, establishing expert completion time as a validity criterion for problem difficulty (Laurent et al., 2024). This aligns with the broader recognition in cognitive science that expert performance emerges from deliberate practice on challenging problems rather than the rapid execution of routine tasks (Ericsson et al., 1993; Ericsson et al., 2018). However, reliance on current expert capabilities may underestimate the AI potential for novel problem-solving approaches that bypass human cognitive constraints (Eloundou et al., 2023; Korinek & Stiglitz, 2021).

Safety and error detection evaluation represents perhaps the most critical advancement, addressing what Amodei et al. (2016) characterize as the "concrete problems in AI safety." BioLP-bench's finding that state-of-the-art models poorly identify critical protocol errors underscores the gulf between linguistic fluency and genuine experimental competence (Ivanov, 2024). This echoes broader concerns about "automation complacency" (Parasuraman & Manzey, 2010) and "algorithmic aversion paradox" (Dietvorst et al., 2015) where users either overtrust sophisticated systems or reject them entirely after observing errors. The pharmaceutical industry's extensive documentation of AI-caused failures, Exscientia's termination of EXS-21546 after failed Phase I trials and Recursion's discontinued programs despite computational predictions, demonstrates that computational validation alone proves insufficient (Exscientia, 2023; Recursion, 2024). CRISPR-GPT's ethical guardrails exemplify the necessary safety constraints, but also highlight the challenges of comprehensive safety evaluation across diverse experimental contexts (Sandbrink, 2023).

Knowledge synthesis evaluation through frameworks such as Dyport and ScholarQABench addresses what Kitano (2021) terms the 'information overload crisis' in biomedical research, where exponential literature growth exceeds human comprehension capacity. Dyport's importance-weighted metrics, which assess not only accuracy but also potential research impact, represent methodological sophistication beyond traditional link prediction (Tyagin & Safro, 2024). This aligns with scientometric research emphasizing that scientific value derives not from the volume of correct predictions, but from identifying high-impact discoveries (Rzhetsky et al., 2015; Wang & Barabási, 2021). However, retrospective validation using temporal cutoff dates, while rigorous, cannot fully capture prospective discovery values, where paradigm shifts may render current importance metrics obsolete (Kuhn, 1970; Lakatos, 2014).

Tool-augmented workflow evaluation reflects the practical recognition that research capabilities emerge from human-AI tool ecosystems rather than isolated model capabilities (Hutchins, 1995; Hollan et al., 2000). LAB-Bench's documentation that tool augmentation significantly improves performance on database queries and literature searches validates the "tools as cognitive prosthetics" framework from distributed cognition theory (Clark & Chalmers, 1998; Zhang & Norman, 1994). However, current benchmarks evaluate tool use in simplified settings that may not be generalizable to real laboratory environments with unreliable instruments, ambiguous data, and emergent failures (Hoffman et al., 2018).

## 4.2 Methodological Trilemma: Scalability, Validity, and Safety

The methodological paradigms identified, literature understanding, experimental design, and hypothesis generation expose a fundamental trilemma in which scalability, validity, and safety can only hardly be simultaneously maximized using current approaches. Literature understanding benchmarks achieves high scalability through automated metrics and curated datasets, enabling continuous leaderboard updates that drive rapid model improvement (Bowman & Dahl, 2021; Kiela et al., 2021). However, this scalability comes with high validity costs. Meta-analyses of benchmark performance reveal persistent "clever Hans" effects where models exploit dataset artifacts rather than learning intended capabilities (Geirhos et al., 2020; Nguyen et al., 2015). The "Goodhart's Law" phenomenon, when measures become targets, they cease to be good measures, manifests clearly in benchmark gaming behaviors (Strathern, 1997; Thomas & Uminsky, 2020).

Experimental design benchmarks address validity through expert evaluation and real-world execution testing, as exemplified by BioPlanner's successful laboratory implementation (O'Donoghue et al. 2023). This approach resonates with arguments from science and technology studies, emphasizing that validity is derived from performance in practice rather than theoretical correctness (Latour & Woolgar, 2013; Pickering, 2010). However, scalability suffers dramatically. BioLP-bench's model-graded evaluation requires GPT-4o-level scoring capabilities, introducing computational costs estimated at $10-50 per evaluation (Ivanov, 2024; OpenAI, 2024). Moreover, reliance on evaluator model quality creates recursive dependency loops, where benchmark validity depends on the presumed capabilities of systems being evaluated (Perez et al., 2022; Chakraborty et al., 2017).

Safety considerations exacerbate this trilemma. Systems capable of designing experiments with dual-use potential, synthetic biology, gain-of-function research, and chemical weapon precursors require safety evaluations before public release (Sandbrink, 2023; Gopal et al., 2023). However, a comprehensive safety evaluation across diverse experimental contexts proves computationally intractable: the combinatorial explosion of possible experiments, reagents, and conditions exceeds feasible testing resources (Anthropic, 2023; UK AI Safety Institute, 2024). The ethical guardrails of CRISPR-GPT demonstrate rule-based approaches, but cannot anticipate novel misuse pathways in rapidly evolving scientific domains (Urbina et al., 2022).

This trilemma explains the two-tier ecosystem in which pharmaceutical companies, Insilico Medicine, Recursion, BenevolentAI, avoid public benchmarks for proprietary validation. Recursion's 2.2 million experiments per week, and BenevolentAI's 48-hour baricitinib identification represents validated capabilities that no benchmark adequately assesses (Recursion, 2025; Richardson et al., 2020). The resulting information asymmetries challenge scientific progress: researchers cannot compare systems objectively, replicate industrial results, or identify transferable insights (Stodden 2010; Wallach et al. 2018). This parallels the broader tensions in AI development between open science norms and competitive/safety imperatives (Liesenfeld et al., 2023; Seger et al., 2023).

## 4.3 Critical Gaps and Methodological Blind Spots

Four major gaps emerged with significant implications for field development. Research proposal evaluation remains unaddressed despite proposals requiring synthesis of literature review, hypothesis generation, experimental design, feasibility assessment, and communication, specifically, the multi-competency integration distinguishing research assistants from narrow tools (Latour, 1987; Collins & Evans, 2019). While tools for proposal writing assistance proliferate, no standardized frameworks have systematically evaluated proposal quality. This gap proves particularly consequential, given that proposal evaluation represents a key bottleneck in research funding (Gallo et al., 2014; Graves et al., 2011) and resource allocation decisions determine research trajectories (Azoulay et al., 2019). The absence of proposal benchmarks may reflect inherent difficulties: proposal quality correlates only weakly with research outcomes (Graves et al., 2011); evaluation involves subjective judgments resistant to formalization (Derrick & Samuel, 2016); and ground truth requires years of prospective tracking (Boudreau et al., 2016). An overview of the gaps identified is shown in Figure 2.

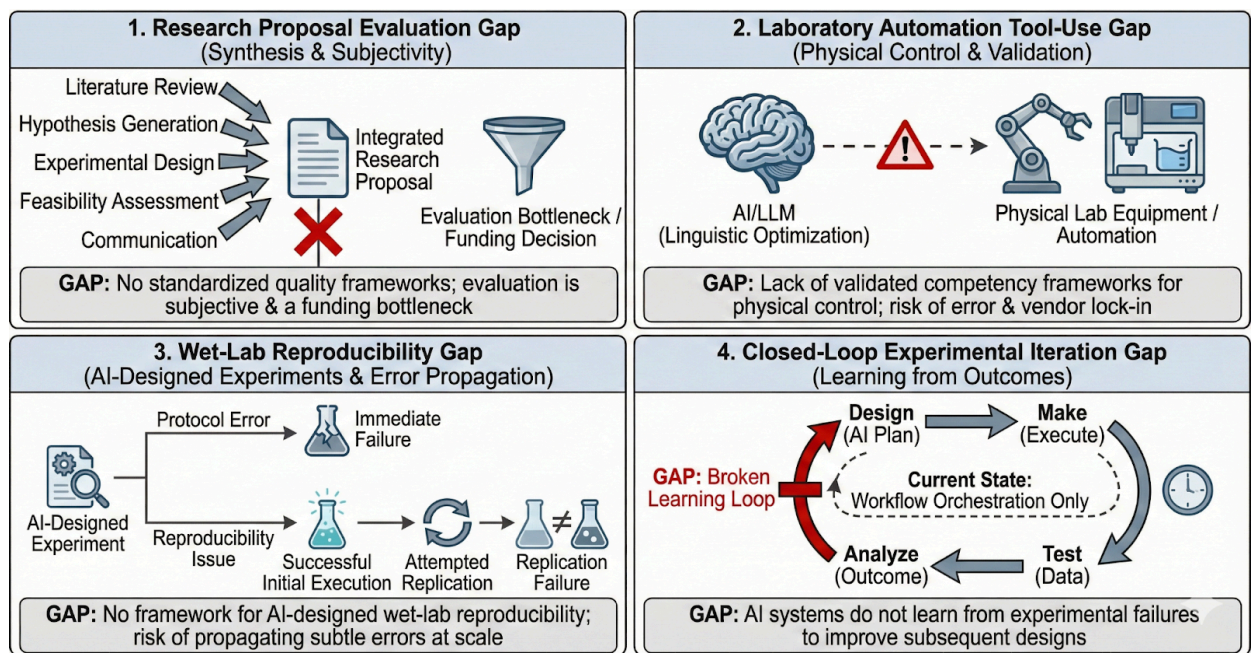

*Figure 2. Four critical deployment gaps beyond benchmarking scope. Panel 1: No standardized frameworks for evaluating integrated research proposals, creating subjective funding bottlenecks. Panel 2: Lack of validated competency frameworks for AI control of physical laboratory equipment. Panel 3: No frameworks for ensuring reproducibility of AI-designed wet-lab experiments or preventing error propagation. Panel 4: AI systems cannot learn from experimental outcomes to improve subsequent designs, breaking closed-loop discovery cycles.*

Laboratory automation tool-use evaluation lags despite the rapid adoption documented in industry reports: 7-8% annual compound growth, extensive AI/ML integration, and productivity gains through platforms such as Emerald Cloud Lab (CMU, 2021; MarketsandMarkets, 2024). The absence of standardized evaluation creates risks of deploying systems optimized for linguistic tasks to control physical equipment without validated competency frameworks, a concern well-established in robotics and human-robot interaction research (Billard & Kragic, 2019; Ajoudani et al., 2018). Moreover, the lack of universal interfaces perpetuates vendor lock-in and limits system interoperability, mirroring historical patterns in laboratory information management systems (Oakley et al., 2025). Carnegie Mellon's Cloud Lab partnership demonstrates academic interest, but has not yet yielded public benchmarks (CMU, 2021).

Wet-lab reproducibility frameworks remain underdeveloped despite reproducibility crises in both biological research (Baker, 2016; Errington et al., 2021) and AI-generated scientific output (Kapoor & Narayanan, 2023; Gibney, 2022). RENOIR addresses ML reproducibility in biomedical sciences, and CORE-Bench evaluates computational reproducibility; however, no framework assesses the reproducibility of AI-designed wet-lab experiments (Barberis et al. 2024; Siegel et al., 2024). Protocol errors (addressed by BioLP-bench) and reproducibility issues represent distinct challenges: errors cause immediate failure, whereas reproducibility problems emerge through attempted replication. This distinction is important because laboratory automation enables rapid experimental throughput, creating the risk of propagating subtle errors on an unprecedented scale (Freedman et al., 2015; Begley & Ioannidis, 2015).

Closed-loop experimental iteration evaluation remains critically underdeveloped, despite its centrality to authentic research practices. Drug discovery fundamentally operates through iterative design-make-test-analyze cycles, where experimental outcomes inform subsequent design decisions (Seal et al., 2025); however, evaluation frameworks lag behind this reality. While BioDiscoveryAgent demonstrates an iterative experimental design for genetic perturbations, achieving a 21% improvement over Bayesian optimization through multi-cycle hypothesis testing (Roohani et al., 2024), such closed-loop evaluations remain exceptional. Current benchmarks, including the proposed framework, predominantly assess whether AI systems help researchers plan experiments but not whether they learn from experimental outcomes to improve subsequent designs. Real-world agentic systems achieve dramatic workflow compression, with literature analysis accelerating from weeks to hours and protocol generation showing >400× speedups; however, these accomplishments reflect orchestration efficiency rather than experimental learning from failures (Seal et al., 2025). This gap is particularly consequential because laboratory automation platforms enable rapid experimental throughput. For example, Recursion Pharmaceuticals executes 2.2 million experiments weekly (Recursion, 2025), creating data volumes in which autonomous iteration can dramatically accelerate discovery. The distinction between conversation-to-proposal evaluation and full experimental loop assessment mirrors the difference between research-planning assistants and genuine discovery partners. For example, when flow cytometry results contradict predictions, preliminary assays fail unexpectedly, or synthesis attempts produce unexpected byproducts, can AI systems help researchers reformulate hypotheses coherently and suggest systematic parameter adjustments? Closed-loop evaluation requires actual experimental execution with substantial time investments but represents an essential capability as AI transitions toward autonomous laboratory operations (CMU, 2021; MarketsandMarkets, 2024).

## 4.4 The Workflow Integration Gap: From Task Evaluation to Collaborative Assessment

Beyond domain-specific coverage gaps, a fundamental paradigm limitation characterizes current benchmarking practices: all 14 reviewed benchmarks evaluate AI systems on isolated tasks or component capabilities, whereas authentic research collaboration requires integrated workflows spanning multiple sessions, adaptive dialogue, and contextual memory (Cetina, 1999; Latour & Woolgar, 2013). Even benchmarks explicitly designed for complex reasoning evaluate task completion rather than the collaborative process quality. This paradigm reflects the historical development of AI benchmarks from the NLP and machine learning communities, where component-level evaluation enables rigorous comparison and rapid iteration (Dodge et al., 2021; Ethayarajh & Jurafsky, 2020). However, research assistance differs fundamentally from isolated task execution in that component-level benchmarks cannot capture the collaborative process.

Research workflows inherently require capabilities that are absent from current evaluation frameworks. Scientists rarely complete projects in single sessions; instead, research progresses episodically over days, weeks, or months, with irregular interaction patterns (Cetina, 1999; Latour & Woolgar, 2013). Budget constraints emerge mid-project and require adaptive replanning that maintains consistency with prior decisions. Experimental results necessitate hypothesis refinement through iterative dialogue, where scientists challenge initial interpretations, and collaborators must incorporate critique gracefully rather than defending conclusions (Collins, 1992; Dunbar, 1995). Current benchmarks evaluate data analysis quality, hypothesis validity, protocol appropriateness, and proposal writing as separate tasks. None assess whether systems maintain context across sessions, propagate constraints across decision points, or adapt to iterative critique. Excellence in component tasks does not guarantee successful collaboration; an agent could score highly on LAB-Bench protocol planning, Dyport hypothesis generation, and BioML-bench data analysis while catastrophically failing integrated research support by forgetting constraints across sessions, defending erroneous interpretations despite corrections, or requiring complete context repetition after temporal gaps.

This workflow integration gap manifests across four critical dimensions entirely absent from current benchmarks. Dialogue quality assessment would evaluate whether agents ask appropriate clarifying questions before committing to interpretations, provide explanations calibrated to researcher expertise, and incorporate corrections gracefully rather than defending errors. Workflow orchestration measurement would assess whether later research stages appropriately reflect earlier decisions and constraints, whether budget limitations discussed in one session propagate to recommendations in subsequent sessions, and whether agents help researchers progress logically through interconnected workflow stages. Session continuity evaluation would test whether systems maintain context across temporal gaps ranging from hours to weeks, distinguishing persistent information (budget constraints, key decisions) from transient details (temporary file names, superseded analyses). Researcher experience measurement would assess trust calibration (whether researchers develop appropriate confidence in agent outputs), cognitive load, satisfaction, and learning support. These dimensions determine whether AI systems function as effective research collaborators rather than merely competent task executors. The absence of workflow-oriented benchmarks creates risks analogous to those in human-robot collaboration research: optimizing component capabilities without evaluating integrated performance yields systems that excel in controlled settings but fail in authentic deployment contexts (Hoffman & Breazeal, 2004; Nikolaidis et al., 2017). This paradigm gap motivates the development of process-oriented evaluation frameworks specifically designed to assess AI research co-pilots on the collaborative dimensions that determine practical utility in extended research workflows. A summary of the workflow integration is shown in Figure 3.

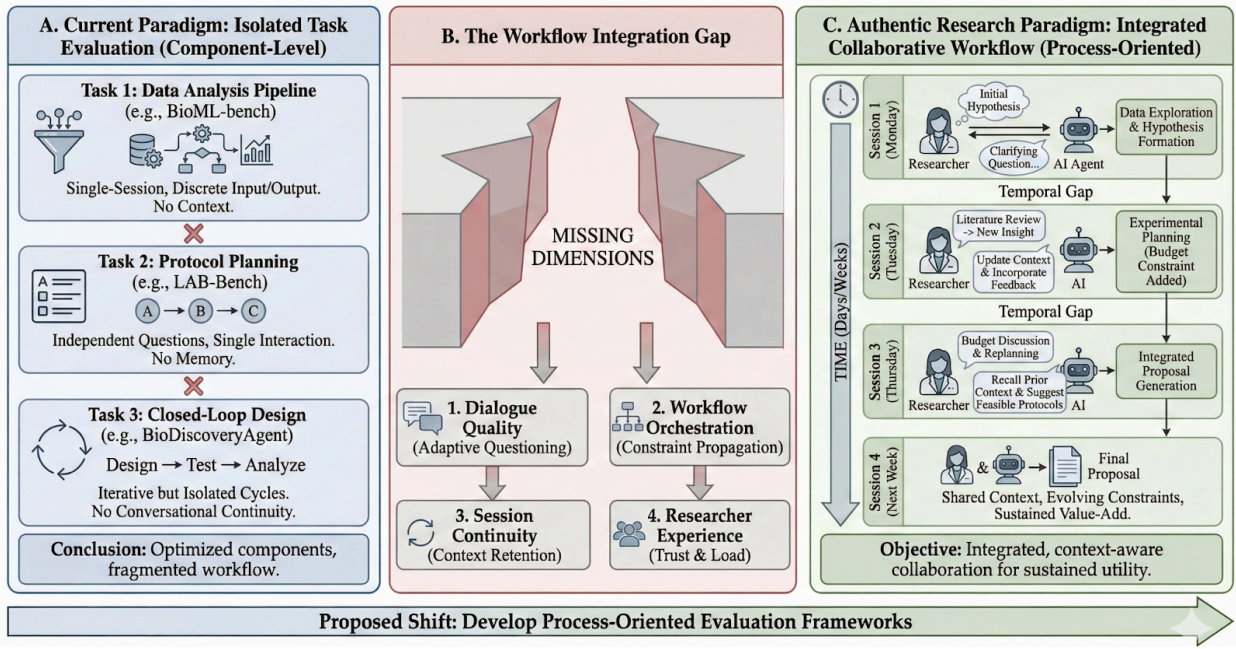

*Figure 3. The paradigm gap in current AI benchmarking approaches.* Panel A: Current benchmarks evaluate isolated tasks without contextual memory or conversational continuity. Panel B: Four critical workflow integration dimensions are missing from evaluation frameworks. Panel C: Authentic research requires multi-session collaboration with temporal gaps, shared context, constraint propagation, and adaptive dialogue across integrated workflows spanning days or weeks.

# 5. The Workflow Integration Gap: Towards Conversational Research Co-Pilot Benchmarking

To bridge the gap between component-level evaluation and authentic research collaboration, this section develops a process-oriented benchmarking framework that assesses AI systems across integrated, multisession research workflows.

## 5.1 The Disconnect Between Component Testing and Research Workflows

The benchmarks examined in this review evaluate AI systems through component-level assessment: literature understanding (BLUE, BLURB, and BioASQ), protocol planning (LAB-Bench and BioPlanner), hypothesis generation (Dyport), and experimental design (CRISPR-GPT). While this approach enables rigorous and scalable evaluation of specific capabilities, it fundamentally misaligns with how researchers use AI research assistants in practice. Real scientific research unfolds through integrated, conversational workflows spanning multiple sessions, requiring systems to orchestrate diverse capabilities while maintaining dialogue coherence and adapting to evolving constraints (Latour 1987; Collins & Evans 2019). BioASQ exemplifies this single-session evaluation paradigm: experts formulate questions reflecting authentic information needs, systems retrieve relevant documents and generate answers, and assessments occur within discrete, time-bounded interactions (Krithara et al., 2023). While this approach enables rigorous comparison across systems and has successfully advanced biomedical question-answering capabilities, it fundamentally cannot evaluate whether systems maintain coherent context when researchers return after temporal gaps or adapt recommendations when constraints emerge across multiple sessions (Krithara et al., 2023; Wu et al., 2024).

This disconnection manifests in three dimensions. First, temporal integration: research projects evolve over days, weeks, or months, with researchers returning to refine hypotheses, adjusting designs based on new information, or pivoting directions after preliminary results. Current benchmarks evaluate single-session task completion, ignoring whether systems can maintain a coherent context across temporal gaps that characterize real research. Constraint negotiation: practical research involves continuous trade-offs, such as budget limitations, equipment availability, ethical considerations, institutional review requirements, and collaborator expertise. Benchmarks typically present idealized scenarios without these messy realities, failing to assess whether systems can adapt recommendations when researchers introduce real-world constraints mid-conversation. Third, conversational quality: The process of arriving at solutions matters as much as the solution quality itself. A system might generate an excellent experimental design, but frustrates researchers with unclear explanations, failure to ask clarifying questions, or inability to incorporate feedback, rendering it practically unusable despite strong benchmark performance (Cai et al., 2019; Bertrand et al., 2022).

The absence of workflow-oriented evaluation creates a validity gap similar to that in human-computer interaction research when usability testing focuses on isolated task completion rather than integrated work practices (Hollan et al., 2000). As Hutchins (1995) demonstrated in studies of distributed cognition, competence emerges from human-tool-environment systems, not individual components in isolation. Analogously, AI research assistance capability manifests through successful collaboration over extended research workflows, not through performance on decontextualized benchmarks. This gap becomes particularly consequential as systems approach deployment: laboratory automation platforms enable productivity increases (CMU, 2021), but unlocking this potential requires AI systems that can guide researchers through complex, multi-stage processes rather than merely executing isolated tasks competently.

## 5.2 Case Study: Benchmarking the Data-to-Proposal Research Journey

To illustrate the workflow integration challenge concretely, consider a realistic scenario that current benchmarks cannot adequately evaluate. An overview of the case study is shown in Figure 4.

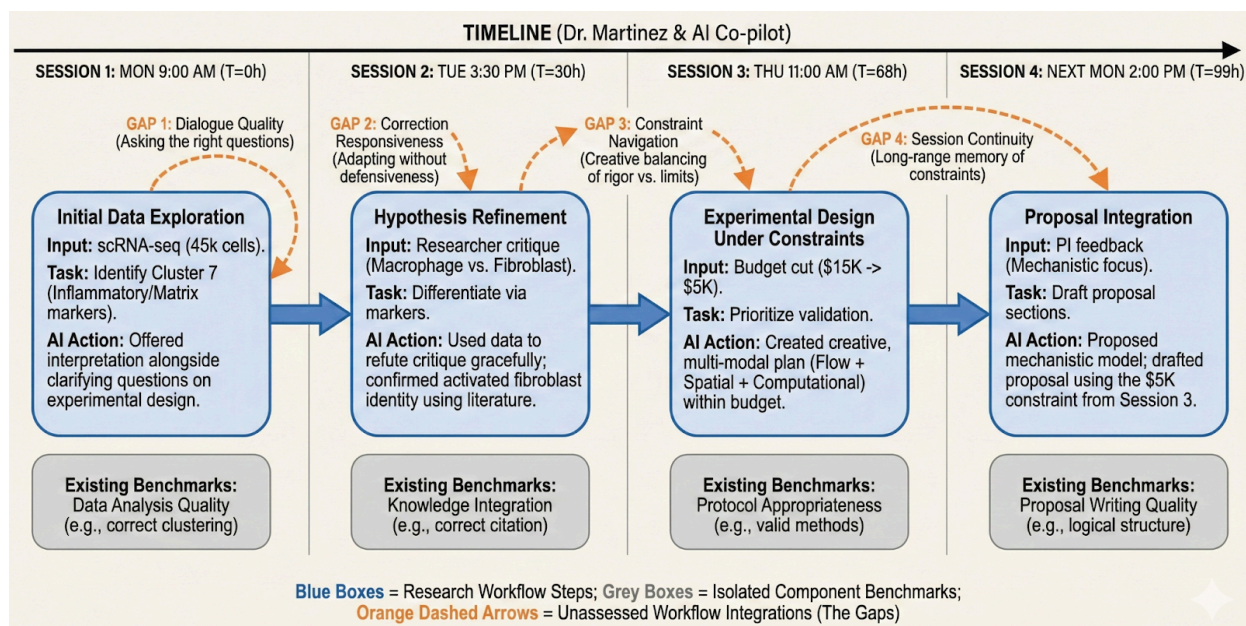

*Figure 4. Case study timeline illustrating workflow integration gaps in current benchmarking.* A realistic research scenario spans four sessions over 99 hours, from initial data exploration through proposal integration. Existing benchmarks (grey boxes) evaluate isolated component capabilities at each stage: data analysis quality, knowledge integration, protocol appropriateness, proposal writing quality. However, they do not assess critical workflow integration dimensions (orange dashed arrows): dialogue quality, correction responsiveness, constraint navigation, and session continuity across temporal gaps.

Dr. Martinez, a cardiovascular biology postdoc, uses an AI research co-pilot to develop a grant proposal:

## 5.2.1 Session 1 (Monday, 9:00 AM): Initial Data Exploration *[Dialogue Quality]*

Dr. Martinez uploads single-cell RNA-sequencing data from failing human hearts, containing 45,000 cells across 12 preliminary clusters identified by standard dimensionality reduction.

**Agent:** *"I've analyzed your scRNA-seq data—12,447 genes detected across 45,238 cells. Cluster 7 is particularly interesting: 2,847 cells showing elevated expression of inflammatory markers (IL6, TNF, CCL2) and extracellular matrix genes (COL1A1, FN1). This doesn't match typical cardiomyocyte, endothelial, or immune cell signatures. Can you tell me about your experimental design and what you expected to find?"*

**Researcher:** *"These are left ventricular tissue samples from 8 patients with heart failure and 3 healthy controls. I expected to see increased fibroblasts and immune infiltration, but cluster 7 is puzzling, it's larger than I anticipated."*

**Agent:** *"Given the marker profile—high inflammatory cytokines plus matrix remodeling genes—cluster 7 might represent activated myofibroblasts or a transitional pro-inflammatory state. This could be central* *to heart failure progression. Would you like me to check recent literature on cardiac fibroblast plasticity, or should we first validate this cluster's identity with additional markers?"*

Current benchmarks evaluate the computational analysis quality demonstrated here: correct gene counts, accurate cluster characterization, appropriate marker identification (BioML-bench, MLAgentBench assess similar analytical capabilities). However, they do not evaluate the dialogue quality: Did the agent appropriately ask clarifying questions about experimental design before committing to an interpretation? Did the agent avoid premature conclusions by offering multiple hypotheses rather than a single definitive answer? These conversational dimensions remain unassessed.

### 5.2.2 Session 2 (Tuesday, 3:30 PM, 30 hours after Session 1): Hypothesis Refinement After Literature Review *[Correction Responsiveness]*

**Researcher**: *"I read the Nature paper you suggested on cardiac fibroblast heterogeneity. I'm now thinking cluster 7 might actually be pro-inflammatory macrophages that share expression signatures with activated fibroblasts. How can we differentiate?"*

**Agent:** *"Excellent critical thinking. Key differentiation markers: macrophages should express CD68, CD163, and MARCO strongly; myofibroblasts express ACTA2, POSTN, and PDGFRA. Looking at your data... cluster 7 shows: CD68 (moderate), ACTA2 (high), PDGFRA (high), POSTN (moderate). This mixed signature actually suggests a distinct activated fibroblast subset rather than macrophages. The inflammatory program might be induced by macrophage-derived signals. Recent work supports this interpretation: Farbehi et al. (2019) identified similar activated fibroblast populations in failing mouse hearts with concurrent inflammatory and matrix-remodeling signatures, and Tallquist's group (Banerjee et al., 2007) showed that cardiac fibroblasts can acquire inflammatory phenotypes in response to IL-1β signaling while maintaining core fibroblast markers. Want to explore the TGF-β or IL-1β signaling pathways that could drive this state?"*

Current benchmarks evaluate the literature retrieval and knowledge integration demonstrated here: accurate citation of relevant papers, appropriate application of prior findings to the current dataset (BioASQ and ScholarQABench assess literature retrieval capabilities; Dyport evaluates hypothesis formation quality). However, they do not evaluate correction responsiveness: Did the agent gracefully incorporate the researcher's macrophage critique, adjusting its hypothesis rather than defending the initial myofibroblast interpretation? Did the agent appropriately credit the researcher's "excellent critical thinking" while providing counterevidence, or did it superficially agree without substance, a common failure mode where LLMs exhibit excessive agreement rather than genuine engagement with critique? These adaptive dialogue behaviors remain unassessed.

### 5.2.3 Session 3 (Thursday, 11:00 AM, 68 hours after Session 1): Experimental Design Under Constraints *[Workflow Orchestration]*

**Researcher**: *"Lab meeting update: PI allocated $5,000 for validation experiments this quarter, not the $15K I requested. We need to prioritize."*

**Agent**: *"Understood. Original protocol had 10-marker flow cytometry panel + bulk RNA-seq validation + immunofluorescence. With budget constraints, here's prioritized approach: (1) Validate cluster 7 identity with 4-marker flow panel (PDGFRA, ACTA2, IL6, COL1A1)—estimated $2,000; (2) Use existing IF antibodies for spatial validation—$800; (3) Defer bulk RNA-seq. This covers essential validation while preserving $2,200 for unexpected costs. Does your core facility offer discounted rates for multi-sample flow cytometry?"*

**Researcher**: *"Yes, 20% discount for 10+ samples. But I'm concerned, will 4 markers be convincing enough for reviewers?"*

**Agent**: *"Valid concern. These 4 markers are literature-supported for activated fibroblasts (Farbehi et al., 2019; Banerjee et al., 2007; Solé-Boldo et al., 2020). Supplement with bioinformatic validation: I can generate pseudotime trajectory analysis and pathway enrichment from your existing data at no cost. This combination—flow validation + spatial confirmation + computational support—should satisfy reviewers while staying in budget."*

Current benchmarks evaluate the experimental protocol design demonstrated here: methodological appropriateness of flow cytometry validation, reasonable cost estimates, scientifically valid marker selection (LAB-Bench assesses protocol planning capabilities). However, they do not evaluate constraint navigation: Did the agent appropriately balance scientific rigor with budget limitations? Did it offer creative problem-solving (adding computational validation to compensate for reduced experimental scope) rather than simply scaling down the original plan? These resource-constrained optimization skills remain unassessed.

### 5.3.4 Session 4 (Monday, 2:00 PM, 99 hours after Session 3): Proposal Integration After PI Feedback *[Session Continuity]*

**Researcher**: *"PI wants mechanistic hypotheses, not just descriptive characterization. She specifically asked about therapeutic implications."*

**Agent**: *"Perfect timing to connect findings to mechanism. Based on cluster 7's elevated TGF-β pathway genes and inflammatory markers, I propose this mechanistic model: IL-1β from infiltrating macrophages activates fibroblast inflammatory program → TGF-β autocrine loop → sustained matrix production → fibrosis progression. Therapeutic angle: IL-1 receptor antagonist (anakinra) is FDA-approved for other inflammatory conditions. I'm drafting proposal sections: (1) Mechanistic hypothesis with pathway diagrams; (2) Validation experiments using the $5,000 budget-constrained protocol we designed Thursday; (3) Therapeutic relevance connecting to anakinra. Want to review the mechanism figure first, or should I complete all sections?"*

Current benchmarks evaluate the proposal writing quality demonstrated here: scientifically sound mechanistic hypothesis, appropriate therapeutic connection, logical structure (general scientific writing benchmarks assess similar capabilities). However, they do not evaluate session continuity: Did the agent remember the $5,000 budget constraint from Session 3 (4 days prior) and maintain consistent recommendations? Did it avoid suggesting the original $15K protocol after the multi-day gap? Could it coherently resume the project without requiring complete context repetition? These temporal memory and workflow integration capabilities remain unassessed.

This case study demonstrates that excellence in component tasks does not guarantee successful research collaboration. A component-capable but workflow-deficient agent might: (1) provide scientifically correct analyses but defend initial interpretations rather than incorporating researcher critiques (Session 2), (2) generate valid experimental protocols but fail to adapt creatively under budget constraints (Session 3), or (3) forget the $5,000 budget limitation after the 4-day gap and suggest the original expensive protocol (Session 4). Current benchmarks would rate such an agent highly on isolated capabilities while missing catastrophic workflow failures. Conversely, an agent with moderately lower component scores might excel at dialogue quality, contextual memory, adaptive refinement, and constraint navigation, ultimately providing substantially greater practical value to researchers.

## 5.3 Proposed Framework: Process-Oriented Benchmarking

Existing benchmarks excel at evaluating component capabilities in isolation. An AI system could demonstrate excellence across all these dimensions while failing catastrophically in integrated research collaboration. Consider a researcher analyzing single-cell cardiac data across multiple sessions: the agent might correctly identify unusual cluster characteristics, propose scientifically sound hypotheses, generate valid experimental protocols, and retrieve relevant literature, yet still provide poor research support by forgetting budget constraints between sessions, defending erroneous interpretations despite researcher corrections, or requiring complete context repetition after temporal gaps. Conversely, an agent with moderately lower component scores may excel at workflow orchestration, adaptive refinement, and collaborative value-add, ultimately providing substantially greater practical utility.

To address the identified workflow integration gap, a framework with four evaluation dimensions (including modular extensions) is proposed to assess context maintenance, iterative refinement, constraint propagation, and adaptive dialogue across extended research collaborations. An overview is shown in Figure 5.

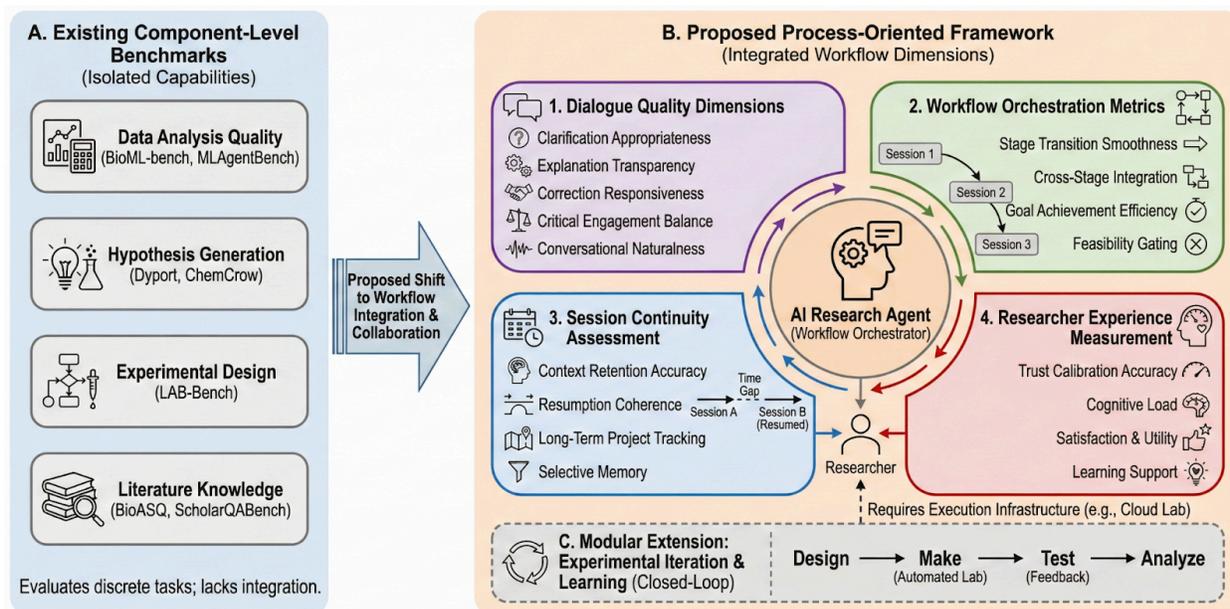

*Figure 5. Paradigm shift from component-level to process-oriented benchmarking.* Existing benchmarks (Panel A) evaluate isolated capabilities without integration assessment. The proposed framework (Panel B) introduces four workflow integration dimensions absent from current benchmarks: Dialogue Quality (clarification, explanation, correction responsiveness, critical engagement, conversational naturalness), Workflow Orchestration (cross-stage integration, constraint consistency), Session Continuity (context retention across temporal gaps), and Researcher Experience (trust calibration, cognitive load, learning support), with optional experimental iteration for laboratory settings.

**Component capabilities assessed by existing benchmarks:**

- Data analysis quality: Computational correctness, statistical validity (BioML-bench, MLAgentBench)
- Hypothesis generation: Scientific plausibility, literature grounding (Dyport, ChemCrow)
- Experimental design: Protocol validity, methodological appropriateness (LAB-Bench)
- Literature knowledge: Citation accuracy, retrieval relevance (BioASQ, ScholarQABench)

**Critical workflow dimensions NOT evaluated by current benchmarks:**

- Dialogue quality: Clarification appropriateness, explanation transparency, correction responsiveness
- Workflow orchestration: Cross-stage integration, constraint consistency, logical progression
- Session continuity: Context retention across temporal gaps, resumption coherence, selective memory
- Researcher experience: Trust calibration accuracy, cognitive load, collaborative value-add

It is important to note that these dimensions represent the concrete work steps of current workflows. Considering the advancement of AI, there are even more dimensions not addressed:

- Novel discovery: Hypothesis generation beyond literature precedent, unconventional experimental designs, non-derivative approaches

- Creative synthesis: Innovative solutions transcending established practices, transformative capabilities unique to human-AI collaboration

### 5.3.1 Dialogue Quality Dimensions

Current benchmarks assess whether agents provide scientifically correct answers but not whether the conversational process supports effective collaboration. When a researcher presents cardiac single-cell data with an unusual cluster, does the agent immediately commit to an interpretation or first ask questions about experimental conditions, prior knowledge, and analytical goals? When proposing that the cluster represents activated myofibroblasts, does the agent explain the marker-based reasoning in terms of researcher expertise or produces either oversimplified summaries or impenetrable technical details? When the researcher challenges this interpretation suggesting macrophage contamination, does the agent incorporate this feedback gracefully or defensively justify its initial conclusion?

Evaluation must assess five core dialogue quality aspects: (1) clarification appropriateness (gathering necessary context before generating recommendations, avoiding premature conclusions based on incomplete information), (2) explanation transparency (providing reasoning that researchers can understand and evaluate, with appropriate detail calibrated to expertise level), (3) correction responsiveness (incorporating researcher feedback gracefully, adjusting conclusions rather than defending errors); (4) critical engagement balance (providing substantive pushback when researcher assumptions are questionable, avoiding excessive agreement that validates flawed reasoning), and (5) conversational naturalness (maintaining coherent dialogue flow, avoiding robotic or repetitive patterns that increase cognitive load).

Measurement approaches combine expert ratings using established dialogue quality frameworks (Deriu et al., 2021; Mehri & Eskenazi, 2020) with turn-level annotations capturing specific dialogue acts, including questions asked, explanations provided, and corrections handled (Bunt et al., 2010; Jurafsky & Shriberg, 1997). The ISO 24617-2 standard for dialogue act annotation provides a robust foundation for the systematic coding of communicative functions across multidimensional dialogue contexts (Bunt et al., 2010). Conversational stress testing scenarios that introduce ambiguous data, conflicting constraints, or simulated researcher confusion reveal agent robustness beyond ideal-case interactions, following established practices in dialogue system evaluation (Jang et al., 2023). Expert annotators assess whether clarification requests appropriately precede commitments, whether explanations match researchers' expertise signals, and whether corrections receive uptake rather than resistance.

### 5.3.2 Workflow Orchestration Metrics

Existing benchmarks evaluate discrete research tasks without assessing the progression through interconnected workflow stages. A cardiac research project might involve Session 1 (exploratory data analysis identifying unusual clusters), Session 2 (hypothesis refinement comparing myofibroblast and macrophage interpretations), Session 3 (budget constraint discussion limiting validation experiments), and Session 4 (final proposal integration). The current benchmarks would evaluate data analysis quality, hypothesis validity, experimental design appropriateness, and proposal writing separately. However, workflow orchestration assesses whether the later stages appropriately reflect earlier decisions. When finalizing the proposal in Session 4, does the agent remember the budget constraint from Session 3 and maintain consistent recommendations or suggest expensive validation experiments incompatible with the stated limitations? Does the experimental design support the refined hypothesis from Session 2 after incorporating macrophage contamination concerns?

Evaluation must assess (1) stage transition smoothness (moving from data exploration to hypothesis formation to experimental design without artificial discontinuities or information loss), (2) cross-stage integration (ensuring later stages reflect constraints and decisions from earlier stages), (3) goal achievement efficiency (minimizing unnecessary conversational turns while maintaining quality), and (4) feasibility gating (identifying unworkable paths early rather than investing effort in infeasible directions).

The measurement combines automated analysis, including conversation turn counts and backtracking frequency, with an expert assessment of workflow coherence (Asai et al., 2024). Visual analytics techniques, including conversation graph representations (nodes representing workflow stages and edges showing information flow patterns), enable structural analysis of workflow quality (Park et al., 2023). Expert evaluators trace specific constraints (budget $5,000, 3-month timeline, and specific equipment availability) through workflow stages, assessing whether recommendations remain internally consistent. Process mining methodologies in business process management offer relevant analytical frameworks for discovering, monitoring, and improving workflow execution patterns (van der Aalst, 2016). A comparison with expert-performed research workflows provided empirical baselines for efficiency and integration quality.

### 5.3.3 Session Continuity Assessment

The evaluated benchmarks operate within single sessions, whereas authentic research collaboration spans days, weeks, or months with irregular interaction patterns. When a researcher returns to the cardiac project after a week-long conference absence, can the agent coherently resume discussion of the myofibroblast hypothesis without requiring complete context repetition? Does it maintain awareness that flow cytometry validation has been constrained by budget limitations or suggest approaches already rejected? Does it remember that the reference by Farbehi et al. (2019) proved particularly relevant and should be emphasized in the proposal? Selective memory becomes critical: the agent must distinguish persistent context (budget constraints, experimental conditions, and key decisions) from transient details (temporary file names and intermediate analysis artifacts superseded by later work).

Evaluation must assess (1) context retention accuracy (correctly recalling previous discussions, decisions, and constraints after varied temporal intervals), (2) resumption coherence (smoothly continuing conversations after gaps without requiring complete context re-establishment), (3) long-term project tracking (maintaining awareness of overall project goals while handling session-specific tasks), and (4) selective memory (distinguishing important persistent context from transient details).

The measurement employs multi-session scenarios with systematically varied gap durations (1 h, 1 d, 1 week, 1 month) incorporating context comprehension probes to test whether the agent retained critical information (Jang et al., 2023; Xu et al., 2021). Recent work on long-term conversational memory provides methodological foundations, including evaluations across dozens of dialogue sessions spanning thousands of turns (Maharana et al., 2024; Kim et al., 2024). Context comprehension probes included direct questions ("What budget constraints did we discuss?"), implicit references ("Given our previous discussion about equipment limitations..."), and consistency checks (detecting contradictions with earlier decisions). Memory decay analysis reveals how context retention degrades over temporal intervals, indicating system design requirements for persistent memory management (Wu et al., 2024). Longitudinal tracking of authentic research projects provides ecological validity, although implementation costs exceed those of controlled experimental designs.

### 5.3.4 Researcher Experience Measurement

The current benchmarks assess objective task performance without evaluating the subjective experience that determines sustained adoption. A researcher collaborating with an agent on cardiac analysis might receive scientifically correct outputs while developing inappropriate trust calibration (accepting suggestions requiring verification or exhaustively checking obvious conclusions), experiencing high cognitive load from confusing interactions, feeling uncertain about agent reasoning, or failing to learn relevant concepts. These experiential dimensions directly affect whether researchers incorporate AI systems into authentic workflows.

Evaluation must assess: (1) trust calibration (researchers' ability to accurately judge when agent suggestions require independent verification, avoiding both over-reliance and under-utilization), (2) cognitive load (mental effort required to collaborate effectively with the agent), (3) satisfaction and perceived utility (willingness to incorporate agent outputs in actual research), and (4) learning support (whether interaction enhanced researchers' understanding of concepts, methods, or their own data).

Trust calibration is a critical metric that assesses whether researcher confidence in agent outputs aligns appropriately with the actual system reliability (Yin et al., 2019; Zhang et al., 2020). Calibration accuracy requires comparing researchers' confidence ratings against independent expert verification of agent outputs across multiple interaction scenarios. When the agent identifies activated myofibroblasts in cluster 7, does the researcher appropriately recognize that this requires validation (rather than acceptance uncritically), while accepting straightforward marker-based cluster annotation without excessive verification? Cognitive workload measurement employs validated instruments, including the NASA Task Load Index (Hart & Staveland, 1988), although recent methodological reviews have highlighted important limitations that require careful interpretation (Babaei et al., 2025). System usability assessment utilizes a System Usability Scale, providing a standardized evaluation of perceived usability across diverse systems (Brooke, 1996; Bangor et al., 2008). Post-interaction interviews can reveal subjective experience nuances that quantitative metrics cannot capture, including whether researchers felt the agent helped them think more clearly, learn relevant concepts, or arrive at insights they might not have reached independently. Longitudinal adoption tracking (do researchers continue using the system after initial exposure, and in what contexts?) provides strong valid signals for practical applications. Comparison with human researcher baselines establishes whether agents provide measurable advantages in task completion time, solution quality, or coverage.

### 5.3.5 Modular Extension: Experimental Iteration and Learning

While the four core dimensions address critical gaps in conversational workflow evaluation, authentic research co-pilots must ultimately learn from the experimental outcomes. This fifth dimension, Experimental Iteration and Learning, assesses whether agents can interpret unexpected results, troubleshoot failures, and systematically refine approaches based on empirical feedback from the design-make-test-analyze cycle (Seal et al., 2025). However, implementing this dimension requires actual experimental execution infrastructure (laboratory automation platforms, synthesis capabilities, and assay systems) with substantial time and cost investments that exceed typical benchmarking feasibility. Therefore, this dimension should be treated as modular rather than a prerequisite. The four core dimensions evaluate essential conversational capabilities applicable across all research contexts, whereas experimental iteration assessment selectively applies to scenarios with available automation infrastructure. Institutions with high-throughput experimental capabilities (cloud labs, automated synthesis platforms) can incorporate closed-loop evaluations, as demonstrated by BioDiscoveryAgent for genetic perturbations (Roohani et al., 2024). Alternatively, computational "experiments" such as code execution, simulation studies, and data analysis pipelines offer lighter-weight testbeds where "experimental outcomes" arrive within seconds, enabling the evaluation of iterative learning without wet-lab requirements. Future standardization efforts should establish protocols for closed-loop evaluation in drug discovery, protocol optimization, and automated data analysis workflows, where rapid iteration cycles prove feasible.

## 5.4 Implementation Considerations

Translating this framework into operational benchmarks faces several methodological challenges. Ground-truth ambiguity poses the primary obstacle: unlike isolated tasks with correct answers, research workflows admit multiple valid paths and outcomes. A researcher might legitimately prioritize comprehensive validation over budget constraints or emphasize mechanisms over phenomenology; neither choice is objectively superior. This necessitates evaluation approaches that recognize valid diversity while identifying clear failures (inconsistent recommendations, context loss, and poor explanations).

Translating this framework into operational benchmarks confronts fundamental methodological challenges inherent in evaluating open-ended conversational systems. Unlike isolated tasks with objectively correct answers, research workflows admit multiple valid approaches and outcomes; a researcher might legitimately prioritize comprehensive validation over budget constraints or emphasize mechanistic understanding over phenomenological descriptions, with neither choice being inherently superior. This ground-truth ambiguity, widely recognized in evaluating open-ended dialogue and creative tasks (Liu et al., 2016), necessitates evaluation frameworks that recognize valid diversity while identifying clear failures such as inconsistent recommendations, context loss, or inadequate explanations. The substantial costs of expert human evaluation present an additional constraint, as domain experts must invest time that could otherwise advance actual research, precisely the resource that AI systems aim to preserve (Deriu et al., 2021; Dinan et al., 2018). Addressing these dual challenges requires tiered evaluation architectures that balance scalability with validity: automated metrics (conversation length, backtracking frequency, goal achievement) provide efficient Tier 1 screening, expert quality assessment concentrates on promising systems in Tier 2, and intensive real-researcher studies evaluate only top performers in Tier 3. This funnel approach, drawing on established practices in dialogue system evaluation, concentrates on expensive human evaluations where it delivers maximum discriminative value (Deriu et al., 2021).

Institutional context independence versus specificity presents a fundamental tension in the benchmark design. For a fair comparison across systems, evaluation scenarios must remain institution-neutral, not favoring agents optimized for specific equipment configurations, model organisms, or methodological traditions. However, real research unfolds within specific institutional contexts that profoundly shape feasible approaches (Latour and Woolgar 2013; Cetina 1999). One institution may have cutting-edge single-cell sequencing capabilities, while another relies primarily on flow cytometry; certain laboratories maintain iPSC-derived cell lines while others work exclusively with primary animal tissue; field-specific norms dictate what constitutes sufficient validation, varying substantially across subdisciplines (Collins, 1992); core facility pricing structures, collaborative arrangements, and reagent accessibility create distinct resource landscapes that fundamentally constrain experimental design (Doing, 2008; Shrum et al., 2007); regulatory environments impose different ethics approval processes, animal use protocols, and biosafety requirements (Leonelli, 2019). Rather than eliminating this variation, which would render scenarios unrealistically generic, benchmark design should treat institutional context as an explicit, modular parameter that systems must navigate, reflecting insights from situated action theory that effective technology use requires adaptation to local conditions (Suchman, 1987; Star & Ruhleder, 1996). The ability to highlight relevant contextual information ("What equipment does your institution have?" "Which model organism does your lab use?") and adapt recommendations accordingly represent a crucial capability for practical research assistance, not a methodological confound. Evaluation scenarios should therefore test context-awareness explicitly: when a researcher states "we don't have access to that sequencing platform," can the agent propose methodologically sound alternatives? When told "our ethics committee requires additional controls for that experiment," does it adjust the protocol appropriately? This approach transforms institutional variation from evaluation obstacles into evaluated capability while simultaneously ensuring fair comparison (all systems face the same contextual constraints) while assessing real-world adaptability. Moreover, requiring agents to handle diverse institutional contexts actually unlocks additional valuable features; systems that can flexibly accommodate varying equipment, organisms, and methodologies are more broadly useful than those optimized for idealized scenarios matching no actual institution, analogous to how machine learning models must generalize across dataset shifts and domain variations (Torralba & Efros, 2011; Quionero-Candela et al., 2022). The resulting benchmarks better predict deployment success: an agent performing well across varied institutional contexts demonstrates robustness that component-level benchmarks, evaluated in context-free settings, cannot assess; a lesson learned repeatedly in medical AI, where models trained in one clinical setting often fail when deployed in institutions with different patient populations, imaging equipment, or clinical workflows (Gulshan et al., 2016).

Reproducibility challenges these evaluation difficulties, as conversational workflows introduce variability from both researcher phrasing choices and model sampling stochasticity, and identical scenarios may unfold differently across runs (Biderman et al., 2024; Marie et al., 2021). Language model evaluation increasingly recognizes that managing such stochasticity requires explicit documentation of sampling parameters, multiple conversation samples per scenario, and evaluation metrics that are robust to superficial variation while detecting substantive failures (Biderman et al., 2024). The distinction between acceptable conversation-level variance (different valid paths to solutions) and problematic session-level inconsistencies (contradicting previous statements within sessions) is critical for meaningful assessments. Beyond the evaluation methodology, workflow benchmarks demand infrastructure exceeding typical benchmark requirements: conversational interfaces rather than simple API endpoints, session management capabilities, context storage mechanisms, and potential integration with research tools including statistical software, literature databases, and laboratory information systems. The democratization of benchmarking practices, exemplified by how HuggingFace's open-source infrastructure lowered adoption barriers for NLP evaluation, suggests that reference implementations demonstrating workflow evaluation would substantially accelerate community adoption and enable systematic progress toward conversational research co-pilots.

## 5.5 Path Forward

Realizing process-oriented benchmarking requires the development of proof-of-concept implementations that demonstrate its feasibility and establish its methodological foundations. The critical first steps include creating standardized research scenarios spanning diverse domains and complexity levels, piloting evaluation protocols with practicing researchers to validate assessment dimensions, and building consensus around evaluation rubrics through multi-institution collaboration. Open-source reference implementations, conversational interfaces, evaluation codes, and baseline systems will prove essential for community adoption by lowering technical barriers that might otherwise limit participation in well-resourced institutions. Engaging the researcher community early ensures that evaluation frameworks capture authentic workflow needs rather than researcher assumptions regarding research practice. As AI research assistants transition from laboratory experiments to practical deployment, developing workflow-oriented evaluation represents not merely methodological refinement but a fundamental necessity: systems optimized for component benchmarks may perform impressively on leaderboards while frustrating researchers whose work demands integrated, multi-session collaboration. The framework proposed here provides a foundation for this evolution, but realizing its potential requires a sustained community effort building evaluation infrastructure that assesses AI systems as research partners rather than isolated task executors.

## 5.6 Limitations

This review focuses exclusively on preclinical biomedical research and excludes clinical and translational applications. Furthermore, the streamline search methodology prioritized efficiency over exhaustive comprehensiveness, although the benchmark concentration in 2022-2024 suggests the focused approach captured recent developments effectively. Gray literature searches may have missed non-English publications or unpublished industry evaluations. The abbreviated search strategy and limited database coverage represent inherent rapid review trade-offs, although forward citation tracking and targeted gray literature searches mitigated the risks of missing major benchmarks. Recent benchmarks (2024-2025) lack longitudinal data on community adoption and practical impacts. Most critically, there is insufficient evidence regarding whether benchmark performance predicts real-world research productivity, measured through publications or discoveries (Ethayarajh & Jurafsky, 2020). The proposed framework lacks empirical validation. While the dimensions draw upon established methodologies from dialogue systems research and HCI, their application to AI research co-pilots requires implementation and testing across multiple systems with domain expert assessment and longitudinal adoption tracking.

However, the suggested framework faces substantial implementation challenges. Workflow evaluation requires a fundamentally different infrastructure than component benchmarking: multi-session scenarios with realistic temporal gaps, conversational interfaces supporting extended dialogue, persistent memory systems, and research tool integration. Expert assessment proves resource-intensive: a single multi-session scenario might require 4-6 hours of expert time for design, simulation, and evaluation compared to seconds for automated component metrics. This scalability challenge mirrors the limitations in dialogue system evaluation, where human judgment remains essential despite costs (Deriu et al., 2021). Ground truth ambiguity compound difficulties: Research workflows allow multiple valid approaches without objectively correct answers. Evaluation must distinguish legitimate methodological diversity from clear failures, such as forgetting constraints or contradicting recommendations, requiring sophisticated judgment to resist full automation. Conversational variability from researcher phrasing and model stochasticity necessitates multiple samples per scenario, explicit sampling documentation, and metrics that are robust to superficial variations while detecting substantive failures (Biderman et al., 2024).

Despite these limitations, the workflow integration gap remains a critical blind spot. Component benchmarks effectively compare isolated capabilities, but cannot assess collaborative effectiveness. Practical adoption may follow tiered approaches: automated metrics (conversation length and backtracking frequency) provide efficient screening, expert assessment concentrates on promising systems, and intensive real-researcher studies validate top performers (Deriu et al., 2021). Future work should develop reference implementations analogous to the HuggingFace infrastructure that democratizes NLP benchmarking (Wolf et al., 2020), enabling community-contributed scenarios and standardized metrics. Longitudinal studies tracking whether workflow evaluation scores predict deployment success would provide essential validation. Interdisciplinary collaboration spanning AI researchers, biomedical scientists, cognitive scientists studying research practices, and HCI experts will prove essential for developing evaluation frameworks that capture the authentic complexity of AI human–research partnerships.

# 6. Conclusion

This review synthesized 14 benchmarks for AI systems in preclinical biomedical research, revealing the sophisticated evaluation of component capabilities, including literature understanding, hypothesis generation, and experimental design. However, a critical paradigm gap has emerged: all current benchmarks evaluate isolated tasks, while authentic research collaboration requires integrated workflows spanning multiple sessions with contextual memory, adaptive dialogue, and constraint propagation. Excellence in component tasks does not guarantee effective research partnership. The proposed four-dimensional framework addresses this gap by evaluating dialogue quality, workflow orchestration, session continuity, and researcher experience, dimensions entirely absent from current benchmarks, yet essential for determining whether AI systems function as practical research collaborators. Implementation faces substantial challenges, including resource-intensive expert assessment, ground-truth ambiguity, and reproducibility concerns. Nonetheless, this workflow integration gap represents a critical blind spot as AI systems transition from narrow task executors to research co-pilots. Future work should develop reference implementations, validate framework dimensions through longitudinal deployment studies, and foster interdisciplinary collaboration to establish evaluation standards that capture the authentic complexity of AI human–research partnerships.

**Funding Statement:**
All authors are employed by Bio.xyz C/O MJP Partners AG, Bahnhofstrasse 20, 6300 Zug, Switzerland. This article was written as part of regular work employment. No additional funding was received for this work.

**AI Use Statement:**
The author(s) declare that Generative AI was used in the creation of this manuscript. During the preparation of this manuscript, the authors used Claude Sonnet 4.5 (Anthropic) to assist with grammar correction, spelling, formatting, and reformulation of selected passages for clarity and style. Image generation was assisted by Nano Banana Pro (Google). All content generated through these tools was critically reviewed, edited, and approved by the authors. Rayyan, a systematic review tool with AI-assisted screening features, was used for deduplication and screening, with all decisions subject to human review. The authors take full responsibility for the integrity and accuracy of the final manuscript.# References

Achiam, J., Adler, S., Agarwal, S., Ahmad, L., Akkaya, I., Aleman, F. L., ... & McGrew, B. (2023). Gpt-4 technical report. arXiv preprint arXiv:2303.08774. https://doi.org/10.48550/arXiv.2303.08774

Ajoudani, A., Zanchettin, A. M., Ivaldi, S., Albu-Schäffer, A., Kosuge, K., & Khatib, O. (2018). Progress and prospects of the human–robot collaboration. Autonomous robots, 42(5), 957-975. https://doi.org/10.1007/s10514-017-9677-2

Amodei, D., Olah, C., Steinhardt, J., Christiano, P., Schulman, J., & Mané, D. (2016). Concrete problems in AI safety. arXiv preprint arXiv:1606.06565.

Anthropic. (2023). Anthropic's Responsible Scaling Policy. https://www.anthropic.com/news/anthropics-responsible-scaling-policy

Asai, A., He, J., Shao, R., Shi, W., Singh, A., Chang, J. C., ... & Hajishirzi, H. (2024). Openscholar: Synthesizing scientific literature with retrieval-augmented lms. arXiv preprint arXiv:2411.14199.

Azoulay, P., Graff Zivin, J. S., Li, D., & Sampat, B. N. (2019). Public R&D investments and private-sector patenting: evidence from NIH funding rules. The Review of economic studies, 86(1), 117-152. https://doi.org/10.1093/restud/rdy034

Babaei, E., Dingler, T., Tag, B., & Velloso, E. (2025). Should we use the NASA-TLX in HCI? A review of theoretical and methodological issues around Mental Workload Measurement. International Journal of Human-Computer Studies, 103515. https://doi.org/10.1016/j.ijhcs.2025.103515

Baker, M. (2016). 1,500 scientists lift the lid on reproducibility. https://doi.org/10.1038/533452a36

Carnegie Mellon University. (2021). Carnegie Mellon University and Emerald Cloud Lab to Build World's First University Cloud Lab. https://www.cmu.edu/news/stories/archives/2021/august/first-academic-cloud-lab.html

Chakraborty, S., Tomsett, R., Raghavendra, R., Harborne, D., Alzantot, M., Cerutti, F., ... & Gurram, P. (2017). Interpretability of deep learning models: A survey of results. In 2017 IEEE smartworld, ubiquitous intelligence & computing, advanced & trusted computed, scalable computing & communications, cloud & big data computing, Internet of people and smart city innovation (smartworld/SCALCOM/UIC/ATC/CBDcom/IOP/SCI) (pp. 1-6). IEEE. 10.1109/UIC-ATC.2017.8397411.

Chen, Q., & Deng, C. (2023). Bioinfo-bench: A simple benchmark framework for llm bioinformatics skills evaluation. BioRxiv, 2023-10. bioRxiv. https://doi.org/10.1101/2023.10.18.563023

Clark, A., & Chalmers, D. (1998). The extended mind. analysis, 58(1), 7-19. https://doi.org/10.1093/analys/58.1.7

Clark, J. M., Sanders, S., Carter, M., Honeyman, D., Cleo, G., Auld, Y., ... & Beller, E. (2020). Improving the translation of search strategies using the Polyglot Search Translator: a randomized controlled trial. Journal of the Medical Library Association: JMLA, 108(2), 195. DOI: 10.5195/jmla.2020.834

Cetina, K. K. (1999). Epistemic cultures: How the sciences make knowledge. Harvard University press.

Collins, H., & Evans, R. (2019). Rethinking expertise. University of Chicago press.

Collins, H. (1992). Changing order: Replication and induction in scientific practice. University of Chicago Press.

Derrick, G. E., & Samuel, G. N. (2016). The evaluation scale: Exploring decisions about societal impact in peer review panels. Minerva, 54(1), 75-97. https://doi.org/10.1007/s11024-016-9290-0

Deriu, J., Rodrigo, A., Otegi, A., Echegoyen, G., Rosset, S., Agirre, E., & Cieliebak, M. (2021). Survey on evaluation methods for dialogue systems. Artificial Intelligence Review, 54(1), 755–810. https://doi.org/10.1007/s10462-020-09866-x

Dietvorst, B. J., Simmons, J. P., & Massey, C. (2015). Algorithm aversion: People erroneously avoid algorithms after seeing them err. Journal of Experimental Psychology: General, 144(1), 114–126. https://doi.org/10.1037/xge0000033

Dinan, E., Roller, S., Shuster, K., Fan, A., Auli, M., & Weston, J. (2018). Wizard of wikipedia: Knowledge-powered conversational agents. arXiv preprint arXiv:1811.01241.
38

Gibney, E. (2022). Could machine learning fuel a reproducibility crisis in science? Nature, 608(7922), 250–251. https://doi.org/10.1038/d41586-022-02035-w

Goldsack, T., Scarton, C., Shardlow, M., & Lin, C. (2024). Overview of the biolaysumm 2024 shared task on the lay summarization of biomedical research articles. arXiv preprint arXiv:2408.08566. https://doi.org/10.18653/v1/2024.bionlp-1.10

Gopal, A., Helm-Burger, N., Justen, L., Soice, E. H., Tzeng, T., Jeyapragasan, G., ... & Esvelt, K. M. (2023). Will releasing the weights of future large language models grant widespread access to pandemic agents?. arXiv preprint arXiv:2310.18233.

Graves, N., Barnett, A. G., & Clarke, P. (2011). Funding grant proposals for scientific research: retrospective analysis of scores by members of grant review panel. Bmj, 343. https://doi.org/10.1136/bmj.d4797

Gu, Y., Tinn, R., Cheng, H., Lucas, M., Usuyama, N., Liu, X., ... & Poon, H. (2021). Domain-specific language model pretraining for biomedical natural language processing. ACM Transactions on Computing for Healthcare (HEALTH), 3(1), 1-23.https://doi.org/10.1145/3458754

Gulshan, V., Peng, L., Coram, M., Stumpe, M. C., Wu, D., Narayanaswamy, A., ... & Webster, D. R. (2016). Development and validation of a deep learning algorithm for detection of diabetic retinopathy in retinal fundus photographs. jama, 316(22), 2402-2410. https://doi.org/10.1001/jama.2016.17216

Hart, S. G., & Staveland, L. E. (1988). Development of NASA-TLX (Task Load Index): Results of empirical and theoretical research. In Advances in psychology (Vol. 52, pp. 139-183). North-Holland.

Hoffman, G., & Breazeal, C. (2004). Collaboration in human-robot teams. In AIAA 1st intelligent systems technical conference (p. 6434).

Hoffman, R. R., Mueller, S. T., Klein, G., & Litman, J. (2018). Metrics for explainable AI: Challenges and prospects. arXiv preprint arXiv:1812.04608.

Hollan, J., Hutchins, E., & Kirsh, D. (2000). Distributed cognition: toward a new foundation for human-computer interaction research. ACM Transactions on Computer-Human Interaction (TOCHI), 7(2), 174-196. https://doi.org/10.1145/353485.353487

Hutchins, E. (1995). Cognition in the wild. MIT Press.

Ivanov, I. (2024). BioLP-bench: Measuring understanding of biological lab protocols by large language models. bioRxiv, 2024-08. https://doi.org/10.1101/2024.08.21.608694

Jang, J., Boo, M., & Kim, H. (2023). Conversation chronicles: Towards diverse temporal and relational dynamics in multi-session conversations. arXiv preprint arXiv:2310.13420.
40